\documentclass[12pt]{article}

\usepackage[round]{natbib} 
\usepackage[margin=1.1in]{geometry}
\usepackage{lmodern}
\usepackage{parskip}

\usepackage[utf8]{inputenc} % allow utf-8 input
\usepackage[T1]{fontenc}    % use 8-bit T1 fonts
\usepackage{hyperref}       % hyperlinks
\usepackage{url}            % simple URL typesetting
\usepackage{booktabs}       % professional-quality tables
\usepackage{amsfonts}       % blackboard math symbols
\usepackage{nicefrac}       % compact symbols for 1/2, etc.
\usepackage{microtype}      % microtypography
\usepackage{hyperref}
\usepackage[pdftex]{graphicx}
\usepackage{xspace}
\usepackage{multirow}
\usepackage{subfigure}
\usepackage{enumitem}
\usepackage{algorithm}
\usepackage{algorithmic}
\usepackage{todonotes}
\usepackage{wrapfig}
\usepackage{cancel}

\usepackage{xcolor}

\title{
Reinforcement Learning \\via Fenchel-Rockafellar Duality
}

\author{%
 \begin{tabular}{p{6cm}p{6cm}}
  {\hfil Ofir Nachum}& {\hfil Bo Dai} \\[.1cm]
  {\hfil \texttt{ofirnachum@google.com}}
    & {\hfil\texttt{bodai@google.com}} \\[.2cm]    
    \hfil \normalfont Google Research & \hfil \normalfont Google Research \\[0.1cm]
  \end{tabular} 
}
\date{}

\usepackage{Definitions}

\newcommand{\comment}[1]{}

\def\f{f}
\def\fdiv{D_{\f}}
\def\fstar{\f_*}
\def\E{\mathbb{E}}
\def\R{\mathbb{R}}
\def\qpi{Q^{\pi}}

\def\jprimal{J_{\mathrm{P}}}
\def\jdual{J_{\mathrm{D}}}

\def\mdp{\mathcal{M}}
\def\init{\mu_0}

\def\visitpi{d^\pi}
\def\visitpistar{d^{\pi^*}}

\def\visitrb{d^\Dset}

\DeclareMathOperator{\avgstep}{\rho}

\def\bellman{\mathcal{B}}

\def\Sset{S}
\def\Aset{A}

\def\Dset{\mathcal{D}}

\def\defeq{:=}
\renewcommand{\widehat}{\hat}

\def\xstar{x^*}
\def\ystar{y^*}
\def\pistar{\pi^*}
\def\zetastar{\zeta^*}
\def\rew{R}
\def\bellman{\gamma\cdot \mathcal{P}^{\pi}}
\def\bellmant{\gamma\cdot \mathcal{P}^{\pi}_{*}}
\def\bellmannog{\mathcal{P}^{\pi}}
\def\bellmantnog{\mathcal{P}^{\pi}_{*}}

\def\vbellman{\gamma\cdot\mathcal{T}}
\def\vbellmant{\gamma\cdot\mathcal{T}_{*}}
\def\vbellmannog{\mathcal{T}}
\def\vbellmantnog{\mathcal{T}_{*}}

\def\dkl{D_{\mathrm{KL}}}

\DeclareMathOperator{\softmax}{softmax}
\def\indicator{\delta}
\def\qvar{Q}
\def\dvar{d}
\def\vvar{V}
\def\KL{\mathrm{KL}}
\def\initsamp{\substack{a_0\sim\pi(s_0) \\ s_0\sim\init}}

\def\ie{\emph{i.e.}\xspace}
\def\eg{\emph{e.g.}\xspace}
\def\cite{\citep}

\begin{document}

\maketitle

\setlength{\abovedisplayskip}{2pt}
\setlength{\abovedisplayshortskip}{2pt}
\setlength{\belowdisplayskip}{2pt}
\setlength{\belowdisplayshortskip}{2pt}
\setlength{\jot}{2pt}

\setlength{\floatsep}{1ex}
\setlength{\textfloatsep}{1ex}

\begin{abstract}
\noindent
    We review basic concepts of convex duality, focusing on the very general and supremely useful Fenchel-Rockafellar duality. 
    We summarize how this duality may be applied to a variety of reinforcement learning (RL) settings,
    including policy evaluation or optimization, online or offline learning, and discounted or undiscounted rewards.
    The derivations yield a number of intriguing results, including the ability to perform policy evaluation and on-policy policy gradient with behavior-agnostic offline data and methods to learn a policy via max-likelihood optimization.
    Although many of these results have appeared previously in various forms, we provide a unified treatment and perspective on these results, which we hope will enable researchers to better use and apply the tools of convex duality to make further progress in RL.
\end{abstract}
\section{Introduction}

Reinforcement learning (RL) aims to learn behavior policies to optimize a long-term decision-making process in an environment. 
RL is thus relevant to a variety of real-world applications, such as robotics, patient care, and recommendation systems.
When tackling problems associated with the RL setting, two main difficulties arise: first, the decision-making problem is inherently sequential, with early decisions made by the policy affecting outcomes both near and far in the future; second, the learner's knowledge of the environment is only through sampled experience, \ie, previously sampled trajectories of interactions, while the underlying mechanism governing the dynamics of these trajectories is unknown.

The environment and its underlying dynamics are typically abstracted as a Markov decision process (MDP). This {abstraction} gives rise to the Bellman recurrence, which characterizes the optimal value function and behavior policy through a dynamic programming (DP) view of the RL problem~\citep{bellman1966dynamic}.
Most of the effective existing RL algorithms are rooted in this dynamic programming paradigm, attempting to find approximate fixed-point solutions to the Bellman recurrence, leading to the family of temporal-difference~(TD) algorithms including SARSA~\citep{sutton1996generalization}, $Q$-learning~\citep{watkins1989learning}, and their deep learning variants~\citep{mnih2015human,van2016deep,wang2015dueling}. 
While the TD-based algorithms are powerful, their training can oscillate or even diverge in settings where function approximation is used or the ability to sample additional interactions with the environment is limited~\citep{sutton1998introduction}.

An alternative paradigm for RL is based on linear programming (LP). A number of RL problems, such as policy optimization and policy evaluation, can be expressed as an LP -- \ie, an optimization problem involving a linear objective and linear constraints. 
The LP may then be transformed to a form more amenable to stochastic and large-scale optimization via the tools of LP duality.
Although the LP perspective has existed for decades (\eg, \citet{manne1960linear,denardo1970linear}), it has recently received renewed interest for its potential ability to circumvent the optimization challenges of DP-based approaches in exchange for more mature and well-studied techniques associated with convex optimization~\citep{de2003linear,wang2007dual,chen2016stochastic,wang2017randomized,bas2019faster}.

In this article, we generalize the LP approach and describe a number of \emph{convex} problems relevant to RL -- \ie, formulations of RL problems as a convex objective and linear constraints.
With a convex objective, one must appeal to the more general Fenchel-Rockafellar duality. 
Perhaps the most useful property of this generalization is that, when the original primal problem involves a strictly convex objective (unlike the LP setting), application of Fenchel-Rockafellar duality leads to a dual problem which is \emph{unconstrained}.
We show that the Fenchel-Rockafellar duality and its variants are at the heart of a number of recent RL algorithms, although many of these were originally presented through less generalizable derivations (\eg, the DICE-family of offline RL algorithms:~\citet{nachum2019dualdice,algae,kostrikov2019imitation,zhang2020gendice}).
By providing a unified perspective on these results and the tools and tricks which lead to them, we hope to enable future researchers to better use the techniques of convex duality to make further progress in RL.

Aiming to provide a useful reference for any interested researcher, we begin by reviewing basic knowledge of convex duality (\secref{sec:convex-duality}) and RL (\secref{sec:rl}). 
We then focus on the \emph{discounted} policy evaluation problem in RL (\secref{sec:eval}), originally expressed in an LP form known as the $Q$-LP. We show how the tools of LP and convex duality may be used to derive a variety of useful re-formulations of policy evaluation.
We then continue to show how these same techniques can be applied to the policy optimization problem, starting from either the $Q$-LP (\secref{sec:opt}) or the potentially more streamlined $V$-LP (\secref{sec:vlp}).
We then generalize these algorithms to \emph{undiscounted} settings in~\secref{sec:undiscounted}. 
We conclude  in~\secref{sec:conclusion} with a brief summary and promising future directions.

\section{Convex Duality}\label{sec:convex-duality}
The concept of duality is a basic and powerful tool in optimization and machine learning, especially in the field of convex analysis, allowing a researcher to easily reformulate optimization problems in alternative ways that are potentially more tractable. In this section, we provide a brief overview of a few key convex duality results which will play an important role in the RL algorithms derived in later sections.

A full and detailed introduction to convex analysis is beyond the scope of this report, and so most of our statements will be presented informally (for example, we use $\min$ and $\max$ as opposed to $\inf$ and $\sup$ and we use $\sum_x f(x)$ and $\int f(x)dx$ without ambiguity), although we will strive to properly qualify theoretical claims when appropriate. The curious reader may refer to a number of resources for a more complete and mathematically precise treatment of this subject; \eg, \citet{rockafellar1970convex,boyd2004convex,borwein2010convex,bauschke2012fenchel}.

\subsection{Fenchel Conjugate}
The \emph{Fenchel conjugate} $\fstar$ of a function $f:\Omega\rightarrow \RR$ is defined as
\begin{equation}
    \label{eq:conjugate}
    \fstar(y) \defeq \max_{x\in\Omega} \inner{x}{y} - f(x),
\end{equation}
where $\inner{\cdot}{\cdot}$ denotes the inner product defined on $\Omega$. This function is also referred to as the \emph{convex conjugate} or \emph{Legendre–Fenchel transformation} of $f$.
% Although it is easiest to consider $f$ as a function $\R\to\R$, in general the domain (and range) of $f$ may be an arbitrary Hilbert space (\eg, $\R^n$) and the term $xy$ in~\eqref{eq:conjugate} should be interpreted as an inner product $\langle x, y \rangle$.

\begin{definition}
We say a function $f$ is proper when $\cbr{x\in \Omega: f(x)< \infty}$ is non-empty and $f(x)>-\infty$ for all $x\in \Omega$.
\end{definition}
\begin{definition}
We say a function $f$ is lower semi-continuous when $\cbr{x\in \Omega: f(x)> \alpha}$ is an open set for all $\alpha\in \RR$. 
\end{definition}
For a proper, convex, lower semi-continuous $f$, its conjugate function $\fstar$ is also proper, convex, and lower semi-continuous. Moreover, one has the duality $f_{**} = f$; \ie,
\begin{align}
    f(x) &= \max_{y\in\Omega^*} \inner{x}{y} - \fstar(y),
\end{align}
where $\Omega^*$ denotes the domain of $\fstar$. From now on we will assume any use of $f$ is for a convex function. Furthermore, we will assume any declared convex function is also proper and lower semi-continuous.~\tabref{tab:conjugates} provides a few common functions and their corresponding Fenchel conjugates. 
\begin{table}[h!]
\centering
\setlength{\tabcolsep}{5pt}
\renewcommand{\arraystretch}{1.2}
\begin{tabular}{|c | c | c |} 
 \hline
 Function & Conjugate & Notes \\ [0.5ex] 
 \hline
 $\frac{1}{2}x^2$ & $\frac{1}{2}y^2$ & \\ 
 $\frac{1}{p}|x|^p$ & $\frac{1}{q}|y|^q$ & For $p,q>0$ and $\frac{1}{p} + \frac{1}{q} = 1$. \\
 $\indicator_{\{a\}}(x)$ & $\langle a, y\rangle$ & $\indicator_C(x)$ is 0 if $x\in C$ and $\infty$ otherwise.\\
 $\indicator_{\R_+}(x)$ & $\indicator_{\R_-}(y)$ & $\R_{\pm}\defeq\{x\in\R~|~\pm x\ge0\}$. \\
 $\langle a, x\rangle + b\cdot f(x)$ & $b\cdot\fstar\left(\frac{y-a}{b}\right)$ & \\
 $D_f(x \| p)$ & $\E_{z\sim p}[\fstar(y(z))]$ & For $x:\mathcal{Z}\to\R$ and $p$ a distribution over $\mathcal{Z}$. \\
 $\dkl(x || p)$ & $\log\E_{z\sim p}[\exp{y(z)}]$ & For $x\in\Delta(\mathcal{Z})$, \ie, a normalized distribution over $\mathcal{Z}$. \\ [1ex] 
 \hline
\end{tabular}
\caption{A few common functions and their corresponding Fenchel conjugates.}
\label{tab:conjugates}
\end{table}

\subsubsection{The Indicator Function}
One especially useful function is the indicator function $\indicator_{C}(x)$, which is defined as,
\begin{equation}
    \indicator_{C}(x) \defeq \begin{cases} 
      0 & x\in C \\
      \infty & \text{otherwise}.
   \end{cases}
\end{equation}
If $C$ is a closed, convex set, it is easy to check that $\indicator_{C}$ is proper, convex, and lower semi-continuous. The indicator function can be used as a way of expressing constraints. For example, the optimization problem $\min_{Ax=0} f(x)$ may be alternatively expressed as $\min_x f(x) + \indicator_{\{0\}}(Ax)$. It may be readily shown that the conjugate of $\indicator_{\{a\}}(x)$ is the linear function $\langle a, y\rangle$ and vice-versa.

\subsubsection{$f$-Divergences}
The family of $f$-divergences, also known as Csiszár-Morimoto or Ali-Silvey divergences~\citep{AliSil66}, has been widely applied in many machine learning applications, including variational inference~\citep{WaiJor03}, generative model estimation~\citep{NowCseTom16,dai2019exponential}, imitation learning~\citep{ke2019imitation,ghasemipour2019divergence,kostrikov2019imitation}, and reinforcement learning~\citep{nachum2019dualdice,zhang2020gendice,algae}. 

For a convex function $f$ and a distribution $p$ over some domain $\mathcal{Z}$, the $f$-divergence is defined as,
\begin{equation}\label{eq:f-div}
\fdiv(x \| p) = \E_{z\sim p}\left[\f\left(\frac{x(z)}{p(z)}\right)\right].
\end{equation}
Typically, $f$-divergences are used to measure the discrepancy between two distributions (\ie, $x\in\Delta(\mathcal{Z})$, the simplex over $\mathcal{Z}$, and $\fdiv(x\|p)$ measures the divergence between $x$ and $p$), although the domain of $\fdiv(x\|p)$ may be extended to the set of real-valued functions $x:\mathcal{Z}\to\R$.

The choice of domain of $\fdiv$ is important when considering the Fenchel conjugate of $\fdiv(x\|p)$. If the domain is the set of unrestricted real-valued functions, the conjugate of $\fdiv(x\|p)$ at $y:\mathcal{Z}\to\R$ is, under mild conditions,\footnote{Conditions of the interchangeability principle~\citep{dai2016learning} must be satisfied, and $p$ must have sufficient support over $\mathcal{Z}$.}
\begin{align}
    \label{eq:fdiv-conjugate}
    g(y) &= \max_{x:\mathcal{Z}\to\R} \sum_z x(z)\cdot y(z) - \E_{z\sim p}[f(x(z) / p(z))] \\ 
    &= \E_{z\sim p} \big[\max_{x(z)\in\R} x(z)\cdot y(z) / p(z) - f(x(z) / p(z))\big] \\ &= \E_{z\sim p}[\fstar(y(z))].
\end{align}
On the other hand, if one considers the domain of $\fdiv$ to be $\Delta(\mathcal{Z})$, then one must solve a constrained version of~\eqref{eq:fdiv-conjugate}, which can be difficult depending on the form of $f$.

\paragraph{The $\KL$-Divergence} Of the family of $f$-divergences, the $\KL$-divergence is arguably the most commonly used one, and it is given by,
\begin{equation}
    \dkl(x\| p) \defeq \sum_z x(z) \log \frac{x(z)}{p(z)},
\end{equation}
which is the result of choosing $f(x)=x\log x$ in~\eqref{eq:f-div}. For the $\KL$-divergence, the constrained version of~\eqref{eq:fdiv-conjugate} may be shown\footnote{See Example 3.25 in~\citet{boyd2004convex}.} to yield the conjugate function
\begin{equation}
    g(y) = \log \E_{z\sim p}[\exp y(z)].
\end{equation}
It is no coincidence that the $\log$-average-$\exp$ function (and the closely related $\log$-$\sum$-$\exp$ function) is arguably as ubiquitous in RL as the $\KL$-divergence.

\subsection{Fenchel-Rockafellar Duality}
Fenchel conjugates are indispensable when tackling a variety of optimization problems. %, leading to the dual form which will be much easier to be solved than the original one. 
In this section, we present one of the most general and useful tools associated with Fenchel conjugates, known as the Fenchel-Rockafellar duality~\citep{rockafellar1970convex,borwein2010convex}.

Consider a primal problem given by
\begin{equation}
    \label{eq:jprimal}
    \min_{x\in \Omega} \jprimal(x) \defeq f(x) + g(Ax),
\end{equation}
where $f,g:\Omega\to\RR$ are convex, lower semi-continuous functions and $A$ is a linear operator (\eg, a matrix). 
The dual of this problem is given by
\begin{equation}
    \label{eq:jdual}
    \max_{y\in\Omega^*} \jdual \defeq -\fstar(-A_* y) - g_*(y),
\end{equation}
where we use $A_*$ to denote the adjoint linear operator of $A$; \ie, $A_*$ is the linear operator for which $\langle y, Ax\rangle = \langle A_*y, x\rangle$, for all $x,y$. In the common case of $A$ simply being a real-valued matrix, $A_*$ is the transpose of $A$.

Under mild conditions,\footnote{See Theorem 3.3.5 in~\citet{borwein2010convex}. Informally, the primal problem needs to be feasible; \ie $\jprimal(x)<\infty$ for some $x\in\Omega$.} the dual problem~\eqref{eq:jdual} may be derived from the primal~\eqref{eq:jprimal} via
\begin{eqnarray}
    \label{eq:dual-deriv}
    \min_{x\in \Omega} f(x) + g(Ax) &=& \min_{x\in \Omega}\max_{y\in\Omega^*} f(x) + \inner{y}{Ax} - g_*(y) \nonumber\\
    &=&\max_{y\in\Omega^*} \cbr{\min_{x\in \Omega} f(x) + \inner{y}{Ax}} - g_*(y)\nonumber\\
    &=&\max_{y\in\Omega^*} \cbr{-\max_{x\in \Omega} \inner{-A_* y}{x} - f(x)} - g_*(y)\nonumber\\
    &=& \max_{y\in \Omega^*} -\fstar(-A_*y) - g_*(y).
\end{eqnarray}
Thus, we have the duality,
\begin{equation}
    \min_{x\in \Omega} \jprimal(x) = \max_{y\in\Omega^*} \jdual(y).
\end{equation}
Furthermore, one may show that a solution to the dual $y^*\defeq \arg\max_y \jdual(y)$ can be used to find a solution to the primal. Specifically, if $\fstar'$ is well-defined, then $x^* = \fstar'(-A_* y^*)$ is a solution to the primal. More generally, one can recover $x^*\in \partial\fstar(-A_* y^*)\cap A^{-1} \partial g_*(y^*)$ as the set of all primal solutions.

Of course, in the presence of Fenchel-Rockafellar duality, the label of primal and dual is arbitrary. One can consider~\eqref{eq:jdual} the primal problem and~\eqref{eq:jprimal} its dual, and in our derivations we will use these labels interchangeably. 

\subsubsection{The Lagrangian}
The Fenchel-Rockafellar duality is general enough that it can be used to derive the Lagrangian duality. Consider the constrained optimization problem
\begin{equation}
    \label{eq:orig-lagrange}
    \min_x\,\, f(x) ~~\st~~ Ax \ge b.
\end{equation}
If we consider this problem expressed as $\min_x f(x) + g(x)$ for $g(x)=\delta_{\R_-}(-Ax + b)$, 
% then following the derivation in~\eqref{eq:dual-deriv}, 
its Fenchel-Rockafellar dual is given by
\begin{equation}
    \label{eq:dual-lagrange}
    \max_y \,\, \langle y, b\rangle - \fstar(A_* y) ~~\st~~ y \ge 0.
\end{equation}
By considering $\fstar$ in terms of its Fenchel conjugate (equation~\eqref{eq:conjugate}), 
we may write the problem as
\begin{equation}
    \min_x\max_{y\ge0}\,\,  \langle y, b\rangle - \langle x, A_* y\rangle + f(x).
\end{equation}
Using the fact that $\langle y, Ax\rangle = \langle x, A_* y\rangle$ for any $A$ we may express this as
\begin{equation}
    \min_x\max_{y\ge0} \,\,
    \underbrace{\langle y, b - Ax \rangle + f(x)}_{L(x, y)}.
\end{equation}
The expression $L(x, y)$ is known as the \emph{Lagrangian} of the original problem in~\eqref{eq:orig-lagrange}.
%\begin{equation}
%    L(x,y) \defeq \langle y, b - Ax \rangle + f(x).
%\end{equation}
One may further derive the well-known Lagrange duality:\footnote{See \citet{veinott} (\url{https://web.stanford.edu/class/msande361/handouts/nlpdual.pdf}) for a brief derivation of this fact and~\citet{EkeTem99}[Proposition 2.1] for more general cases.} 
\begin{equation}
    \max_{y\ge0}\min_x\,\, L(x,y) = \min_x \max_{y\ge0}\,\, L(x,y).
\end{equation}
Moreover, the optimal value of the Lagrangian is the optimal value of the original problem~\eqref{eq:orig-lagrange}, and the optimal solutions (equilibrium points) $\xstar,\ystar$ are the solutions to the original primal~\eqref{eq:orig-lagrange} and its dual~\eqref{eq:dual-lagrange}.

\subsubsection{LP Duality}
The Fenchel-Rockafellar duality also generalizes the well-known linear programming (LP) duality. Specifically, if one considers functions $f(x) = \langle c, x \rangle + \indicator_{\R_+}(x)$ and  $g(x) = \indicator_{\{b\}}(x)$ then the primal and dual problems in~\eqref{eq:jprimal} and~\eqref{eq:jdual} correspond to,
\begin{align}
    \label{eq:lp-primal}
    \min_{x\ge0}\,\, \langle c, x \rangle & ~~\st~~ Ax=b, \\
    \label{eq:lp-dual}
    \max_y\,\, -\langle b, y \rangle & ~~\st~~ A_*y\ge-c, 
\end{align}
respectively. By making the switch $y\to-y$, the dual~\eqref{eq:lp-dual} may be equivalently expressed in the more familiar form,
\begin{equation}
    \label{eq:lp-dual2}
    \max_y\,\, \langle b, y \rangle ~~\st~~ A_*y\le c.
\end{equation}
Fenchel-Rockafellar duality thus provides us with the strong LP duality theorem. Namely, if the primal problem~\eqref{eq:lp-primal} is feasible, then its result is the same as that of the dual~\eqref{eq:lp-dual2}.
\section{Reinforcement Learning}\label{sec:rl}
In this work, we will show how the Fenchel-Rockafellar duality (and the LP and Lagrangian dualities) can be applied to solve a number of reinforcement learning (RL) problems. Before we present these algorithms, we use this section as a brief introduction to RL.

\subsection{Markov Decision Process}
In RL, one wishes to learn a behavior policy $\pi$ to interact with an environment in an optimal way, where the typical meaning of `optimal' is with respect to future discounted rewards (feedback) provided by the environment.
The RL environment is commonly abstracted as a Markov decision process (MDP)~\citep{puterman1994markov, sutton1998introduction}, which is specificied by a tuple
$\mdp = \langle \Sset, \Aset, \rew, T, \init, \gamma \rangle$, consisting of a state space, an action space, a reward function, a transition probability function, an initial state distribution, and a discount factor $\gamma\in (0, 1]$, respectively.
The policy $\pi$ is a function $\Sset\to\Delta(\Aset)$. The policy interacts with the environment iteratively, starting with an initial state $s_0 \sim \init$.  At step $t=0,1,\dots$, the policy produces a distribution $\pi(\cdot|s_t)$ over the actions $\Aset$, from which an action $a_t$ is sampled and applied to the environment.  
The environment produces a scalar reward $r_t=\rew(s_t, a_t)$\footnote{For simplicity we consider a deterministic reward function. Stochastic rewards are more typical, although the same derivations are usually applicable in either case.} 
%based on the current state and chosen action 
and subsequently transitions to a new state $s_{t+1} \sim T(s_t, a_t)$. 
%We denote the value of a policy as 
%\begin{equation}\label{eq:policy_value}
%\rho(\pi) = 
%\begin{cases}
%\lim_{T\rightarrow\infty}\frac{1}{T}\EE\sbr{\sum_{t=0}^T r_t} &\quad \gamma = 1\\
%\rbr{1 - \gamma}\EE\sbr{\sum_{t=0}^\infty \gamma^t r_t} &\quad \gamma\in (0, 1),
%\end{cases}
%\end{equation}
%where the expectation is w.r.t. stochasticity in policy, reward, and environment. 

In summary, the RL setting is concerned with a policy which sequentially makes decisions, and the effects of those decisions are observed through a per-step reward feedback and a stochastic, Markovian state transition process. 
For simplicity, we will consider infinite-horizon (non-terminating) environments, which may be extended to finite-horizon environments by considering an extra terminal state which continually loops onto itself with zero reward.

\subsection{Policy Evaluation and Optimization}
The first question one may ask when presented with an MDP $\mdp$ and a policy $\pi$ is, what is the long-term value of $\pi$ when interacting with $\mdp$? The next question might be, what is the optimal policy $\pistar$ maximizing this long-term value? These two questions constitute the policy evaluation and optimization problems, respectively.

To formalize these questions, we consider a discount factor $\gamma\in(0,1)$.\footnote{See Section~\ref{sec:undiscounted} for consideration of $\gamma=1$.}
The value of $\pi$ is defined as the expected per-step reward obtained by following the policy, averaging over time via $\gamma$-discounting:
\begin{equation}
    \avgstep(\pi) \defeq (1-\gamma) \cdot \E\big[\left.\textstyle\sum_{t=0}^\infty \gamma^t \rew(s_t,a_t) ~\right| s_0\sim\init, \forall t, a_t\sim \pi(s_t), s_{t+1}\sim T(s_t, a_t)\big]. \label{eqn:avgstep}
\end{equation}
The \emph{policy evaluation} problem is to estimate this quantity for a given $\pi$, and the \emph{policy optimization} problem is to find $\pistar$ which maximizes $\avgstep(\pi)$, \ie, solve $\pistar \defeq \arg\max_\pi\avgstep(\pi)$. 
If the reward function $\rew$ is independent of the policy $\pi$, there exists an optimal $\pistar$ that is deterministic~\citep{puterman1994markov}. If one adds a policy-dependent regularization to the objective, \eg, by considering entropy-regularized rewards $\tilde{\rew}(s,a)=\rew(s,a) - \log\pi(a|s)$, the optimal policy $\pistar$ could be stochastic.

\subsection{Online vs. Offline RL}
One of the main limitations when approaching either the policy evaluation or policy optimization problems is that one does not have explicit knowledge of the environment; \ie, one does not have explicit knowledge of the functions $\rew,T,\init$. Rather, access to the environment is given in the form of experience $s_t,a_t,r_t,s_{t+1},a_{t+1},r_{t+1},\dots$ gathered via interactions with the environment.
The specific nature of this experience depends on the context of one's considered problem. The most common forms of experience may be generally categorized into \emph{online} and \emph{offline}.

In the \emph{online} setting, experience from the environment may be collected at any point via Monte-Carlo rollouts.
With this type of access to the environment, the policy evaluation and optimization problems may be easily solved. For example, the value of the policy may be estimated by simply averaging the discounted reward of a large number of Monte-Carlo rollouts.
For this reason, online RL research typically strives to find \emph{sample-efficient} algorithms, which find approximate solutions to policy evaluation or optimization with as few interactions with the environment as possible. 

In practice (\eg, consumer web recommendation systems or health applications), interaction with the environment during training is not available at all. More commonly, access to the environment is \emph{offline}. That is, interactions with the environment are limited to a static dataset of (logged) experience $\Dset=\{(s^{(i)},a^{(i)},r^{(i)},s^{(i)\prime})\}_{i=1}^N$, where $(s^{(i)},a^{(i)})\sim\visitrb$ for some unknown distribution $\visitrb$, $r^{(i)}=\rew(s^{(i)},a^{(i)})$, and $s^{(i)\prime}\sim T(s^{(i)},a^{(i)})$.
One also typically assumes access to samples $\mathcal{U}=\{s_{0}^{(i)}\}_{i=1}^M$ from $\init$. 
%The tuples $(s^{(i)},a^{(i)},r^{(i)},s^{(i)\prime})$ are usually referred to as \emph{transitions}.
In this report we will mostly focus on the offline setting, although we will relax it to assume that our offline experience is effectively unlimited ($N,M\to\infty$), and so will write our expectations in terms of $\visitrb$, $T$, and $\init$. Performing the appropriate finite-sample analysis for finite $N,M$ is outside the scope of this report. Even with effectively unlimited experience, the offline setting presents a challenge to RL algorithms, due to the mismatch between the experience distribution given by the offline dataset and the online distribution typically needed for policy evaluation or optimization.

Although $\visitrb$ is a distribution over state-action pairs, we will at times abuse notation and write $(s,a,s')\sim\visitrb$, and this is intended to mean $(s,a)\sim\visitrb,s'\sim T(s,a)$; \ie, as if we are sampling from the (infinite) dataset $\Dset$. 
Moreover, although one does not have explicit access to $\rew$, we will oftentimes write expressions such as $\E_{(s,a)\sim\visitrb}[\rew(s,a)]$, and this is intended to mean $\E_{(s,a,r,s')\sim\Dset}[~r~]$.

We emphasize a subtle difference between the offline setting and what is commonly referred to in the literature as \emph{off-policy} learning. Off-policy algorithms are designed to enable an RL agent to learn from historical samples collected by other policies. However, these algorithms are typically allowed to interact with the environment during training to collect new samples. On the other hand, in the offline setting, one's access to the environment is exclusively via a fixed dataset of experience. In other words, while an offline RL algorithm is necessarily off-policy, an off-policy algorithm is not necessarily offline.

\subsection{$Q$-values and State-Action Visitations}
When evaluating or optimizing policies, both online and offline, the notions of $Q$-values and state-action visitations are useful.
For a policy $\pi$, the Q-values $\qpi(s,a)$ denote the future discounted sum of rewards of following $\pi$ starting at $s,a$: 
\begin{equation*}
    \qpi(s,a) \defeq \E\big[\left.\textstyle\sum_{t=0}^\infty \gamma^t \rew(s_t,a_t) ~\right| s_0=s,a_0=a, \forall t>0, a_t\sim \pi(s_t), s_{t}\sim T(s_{t-1}, a_{t-1})\big]. \label{eqn:q-value}
\end{equation*}
The $Q$-values satisfy the single-step \emph{Bellman} recurrence
\begin{equation}
    \label{eq:bellman-q}
    \qpi(s,a) = \rew(s,a) + \bellman \qpi(s,a),
\end{equation}
where $\bellmannog$ is the policy transition operator,
\begin{equation}
    \label{eq:bellman}
    \bellmannog \qvar(s,a) \defeq \E_{s'\sim T(s,a),a'\sim\pi(s')}[\qvar(s',a')].
\end{equation}

The state-action visitations $\visitpi$ of $\pi$ (also known as occupancies or density) may be defined similarly as,
\begin{equation*}
\visitpi(s,a) \defeq (1 - \gamma) \textstyle\sum_{t=0}^\infty \gamma^t \Pr\left(s_t=s,a_t=a ~|~ s_0\sim\init, \forall t, a_t\sim \pi(s_t), s_{t+1}\sim T(s_t, a_t) \right).
\end{equation*}
That is, the visitation $\visitpi(s,a)$ measures how likely $\pi$ is to encounter $s,a$ when interacting with $\mdp$, averaging these encounters over time via $\gamma$-discounting. The visitations $\visitpi$ constitute a normalized distribution, and this distribution is referred to as the \emph{on-policy} distribution.

Like the $Q$-values, the visitations satisfy the single-step \emph{transpose Bellman} recurrence:
\begin{equation}
    \label{eq:bellman-d}
    \visitpi(s,a) = (1-\gamma)\init(s)\pi(a|s) + \bellmant\visitpi(s,a),
\end{equation}
where $\bellmantnog$ is the transpose (or adjoint) policy transition operator,
\begin{equation}
    \label{eq:bellmant}
    \bellmantnog \dvar(s,a) \defeq \pi(a|s) \sum_{\tilde{s},\tilde{a}} T(s|\tilde{s},\tilde{a})\dvar(\tilde{s},\tilde{a}).
\end{equation}
These recursions simply reflect the conservation of flow (probability mass) of a stationary distribution on a Markov process.
Note that both $\bellmannog$ and $\bellmantnog$ are linear operators and that the transpose policy transition operator $\bellmantnog$ is indeed the mathematical transpose (or adjoint) of $\bellmannog$ in the sense that $\langle y, \bellmannog x\rangle = \langle \bellmantnog y, x\rangle$ for any $x,y$.

Both the $Q$-values and the visitations are useful in RL. For example, the value of a policy may be expressed in two ways:
\begin{equation}
    \label{eq:avgstep-two-ways}
    \avgstep(\pi) = (1-\gamma)\cdot\E_{\initsamp}[\qpi(s_0,a_0)] = \E_{(s,a)\sim\visitpi}[\rew(s,a)].
\end{equation}
Also, when performing policy optimization, the policy gradient theorem~\citep{sutton2000policy} utilizes the $Q$-values and visitations to express the gradient of $\avgstep(\pi)$ as
\begin{equation}
    \label{eq:policy-gradient}
    \frac{\partial}{\partial\pi} \avgstep(\pi) = \E_{(s,a)\sim\visitpi}[\qpi(s,a)\nabla\log\pi(a|s)].
\end{equation}
It is thus standard in most RL algorithms to either have access to $\qpi$ and $\visitpi$ or have some mechanism to estimate these quantities.
Typically, the $Q$-values are estimated by finding the fixed point of the Bellman recurrence~\citep{SutMaeSze08,SutMaePreBhaetal09,scherrer2010should,LiuLiuGhaMahetal15,dai2016learning,du2017stochastic}; \ie, minimizing the (surrogate) squared difference between the LHS and RHS of~\eqref{eq:bellman-q}, potentially with target networks.
For the visitations, it is more typical to assume access to the distribution $\visitpi$ (for example, by simply performing Monte-Carlo rollouts enabled by online access), although instances exist in which $\visitpi$ is approximated by importance-weighting a different (\ie, offline) distribution~\citep{Precup01OP,sutton2014new}.
\section{Policy Evaluation}
\label{sec:eval}

% \Bo{TODO
% \begin{itemize}
% \item discuss the Fenchel duality in GenDICE,
% \item cite ALGAE in policy evalution via Lagrangian of $Q$-LP.
% \end{itemize}
% } 

We now move on to demonstrating applications of Fenchel-Rockafellar duality to RL. We begin by approaching the policy evaluation problem. Although the policy evaluation problem may appear to be simpler or less interesting (it is not!) than the policy optimization problem, in our case the same techniques will be used in either setting. Thus, we will use
this section to provide more detailed derivations of a variety of techniques which will be referenced repeatedly in the following sections.
%In fact, the policy evaluation is as important as policy optimization in practice. In our case, the same techniques will be used in both policy evaluation and policy optimization problems. Thus, we will provide more detailed derivations of a variety of techniques in this section, which will be referenced repeatedly in the following sections.

\subsection{The Linear Programming Form of $Q$}
\label{sec:q-lp}
The equivalent formulations of $\avgstep(\pi)$ in~\eqref{eq:avgstep-two-ways} in terms of either $\qpi$ or $\visitpi$ hint at a duality which is formally given by the following LP characterization of $\avgstep(\pi)$, known as the $Q$-LP:
\begin{align}
    \avgstep(\pi) = \min_{\qvar} & ~(1-\gamma)\cdot\E_{\initsamp}[\qvar(s_0,a_0)] \\
    \st &~ \qvar(s,a) \ge \rew(s,a) + \bellman \qvar(s,a), \\
    &~\forall (s, a)\in \Sset\times\Aset.
\end{align}
The optimal $\qvar^*$ of this LP satisfies $\qvar^*(s,a)=\qpi(s,a)$ for all $s,a$ reachable by $\pi$. %The resulting value of the LP is the value of the policy $\avgstep(\pi)$. 

The dual of this LP provides us with the visitation perspective on policy evaluation:
\begin{align}
    \label{eq:d-obj}
    \avgstep(\pi) = \max_{\dvar\ge0} & ~\sum_{s,a} \dvar(s,a)\cdot\rew(s,a) \\
    \label{eq:d-const}
    \st &~ \dvar(s,a) = (1-\gamma)\init(s)\pi(a|s) + \bellmant \dvar(s,a), \\
    &~\forall s\in \Sset, a\in\Aset.
\end{align}
The optimal $\dvar^*$ of this LP is the state-action visitation $\visitpi$ of $\pi$. It is important to note that this dual LP is over-constrained. The $|\Sset|\times|\Aset|$ equality constraints in~\eqref{eq:d-const} uniquely determine $d$ regardless of the objective in~\eqref{eq:d-obj}. This fact will prove useful in a number of later derivations.

For detailed and complete derivations of these LP representations of $Q^\pi$ and $\visitpi$, please refer to~\citet{algae}.

\subsection{Policy Evaluation via the Lagrangian}
\label{sec:eval-lagrange}
The potentially large number of constraints in either the primal or dual forms of the $Q$-LP introduce a challenge to estimating $\avgstep(\pi)$. 
We may instead derive a more tractable unconstrained optimization perspective on the policy evaluation problem using the Lagrangian of the $Q$-LP:
\begin{multline}
    \label{eq:eval-lagrange}
    \avgstep(\pi) = \min_{\qvar} \max_{\dvar\ge0} ~~(1-\gamma)\cdot\E_{\initsamp}[\qvar(s_0,a_0)]  + \\ \sum_{s,a}\dvar(s,a)\cdot(\rew(s,a) + \bellman \qvar(s,a) - \qvar(s,a)).
\end{multline}
In practical settings where $\abr{\Sset}\times\abr{\Aset}$ is possibly infinite, it is not feasible to optimize the sum in~\eqref{eq:eval-lagrange} over $\Sset\times\Aset$. In an offline setting, where we only have access to a distribution $\visitrb$, we may make a change-of-variables via importance sampling, \ie, $\zeta(s,a)=\frac{\dvar(s,a)}{\visitrb(s,a)}$. If $\visitrb$ has sufficient support or coverage~\citep{sutton2016emphatic}, we may re-write~\eqref{eq:eval-lagrange} as
\begin{multline}
    \label{eq:ope-lagrange}
    \min_{\qvar} \max_{\zeta\ge0} ~L(\qvar,\zeta) \\
    \hspace{-2mm}\defeq~ (1-\gamma)\cdot\E_{\initsamp}[\qvar(s_0,a_0)] +  \E_{(s,a)\sim\visitrb}[\zeta(s,a)\cdot(\rew(s,a) + \bellman \qvar(s,a) - \qvar(s,a))] \\
    \hspace{-2mm}= (1-\gamma)\cdot\E_{\initsamp}[\qvar(s_0,a_0)] +  \E_{\substack{(s,a,s')\sim\visitrb \\ a'\sim\pi(s')}}[\zeta(s,a)\cdot(\rew(s,a) + \gamma\qvar(s',a') - \qvar(s,a))].\hspace{-2mm}
\end{multline}
The optimal $\zetastar$ of this problem satisfies $\zetastar(s,a)=\frac{\visitpi(s,a)}{\visitrb(s,a)}$.
Thus, to estimate $\avgstep(\pi)$, one may optimize this objective with respect to $\qvar,\zeta$ (requiring only access to samples from $\init$, $\visitrb$, and $\pi$) and return $L(\hat{\qvar}^*,\hat{\zeta}^*)$ as the final estimate.

This more practical, offline estimator also has a desirably property, known as the \emph{doubly robust} property~\citep{funk2011doubly,jiang2015doubly,kallus2019double}. Specifically,
\begin{equation}
    L(\qvar,\zetastar) = L(\qvar^*,\zeta) = L(\qvar^*,\zetastar) = \avgstep(\pi).    
\end{equation}
Thus, this estimator is robust to errors in at most one of $\qvar$ and $\zeta$. 

Despite the desirable properties of this estimator, the optimization problem associated with it involves rewards $\rew(s,a)$ and learning $\qpi$-values with respect to these rewards. Learning $\qpi$-values using single-step transitions turns out to be difficult in practice without the use of a number of tricks developed over the years (\eg, target networks, ensembling). Moreover, the bilinear nature of the Lagrangian can lead to instability or poor convergence in optimization~\citep{dai2017boosting,bas2019faster}.
Rather than tackling these various issues head-on, a number of recent works propose an alternative approach, which we describe in the following subsection.

\subsection{Changing the Problem Before Applying Duality}
%\subsection{Several Variants from Duality}
\label{sec:change-the-problem}
% Recall that the LP for $\visitpi$ is over-constrained. One may replace the objective in~\eqref{eq:d-obj} with $\max_\dvar~-h(\dvar)$ for any $h$ and the optimal $\dvar^*$ would still be $\visitpi$.
% Due to the instability in numerical algorithm for the LP, one may introduce the regularization to~\eqref{eq:d-obj}, 
As mentioned in Section~\ref{sec:q-lp}, the dual of the $Q$-LP is over-constrained, in the sense that the $|S|\times|A|$ constraints in~\eqref{eq:d-const} uniquely determine the state-action visitation $d^\pi$. Thus, one may replace the objective in~\eqref{eq:d-obj} with $\max_\dvar\,\,-h(\dvar)$ for some $h$ without affecting the optimal solution $\dvar^*=\visitpi$.
Therefore, the main idea of a number of recent works is to choose an appropriate $h$ so that either the Lagrangian or the Fenchel-Rockafellar dual of this problem is more approachable and potentially avoids the instabilities associated with the original LP. 

Although the problem is changed, the solution is unaffected, and once a solution is found it may be used to provide an estimate of $\avgstep(\pi)$. Specifically, if the problem is re-written in terms of $\zeta(s,a)=\frac{\dvar(s,a)}{\visitrb(s,a)}$, then once the problem is optimized, we can derive an estimate for the value of $\pi$ via the approximate solution $\hat{\zeta}^*$:
\begin{equation}
    \label{eq:avgstep-zeta}
    \hat{\avgstep}(\pi) = \E_{(s,a)\sim\visitrb}[\hat{\zeta}^*(s,a)\cdot \rew(s,a)].
\end{equation}

\subsubsection{Constant Function $h(\dvar)\defeq 0$}
If $h$ is taken to be the trivial function $h(\dvar)\defeq 0$, the offline form of the Lagrangian optimization becomes,
\begin{multline}
    \label{eq:zero-lagrange}
    \min_{\qvar} \max_{\zeta} ~L(\qvar,\zeta) \\
    = (1-\gamma)\cdot\E_{\initsamp}[\qvar(s_0,a_0)] +  \E_{\substack{(s,a,s')\sim\visitrb \\ a'\sim\pi(s')}}[\zeta(s,a)\cdot(\gamma\qvar(s',a') - \qvar(s,a))].
\end{multline}
The optimal solution $\zetastar$ of this problem is $\visitpi / \visitrb$, and once an approximate solution is found, it may be used to estimate $\avgstep(\pi)$ according to~\eqref{eq:avgstep-zeta}.
Unlike the previous form of the Lagrangian in~\eqref{eq:ope-lagrange}, this optimization does not involve learning $Q$-values with respect to environment rewards, and in practice this distinction leads to much better optimization behavior~\citep{uehara2019minimax}.
Still, the Lagrangian is linear in both $\qvar$ and $\zeta$. This can be remedied by choosing a strictly convex form of $h$, for example, by using an $f$-divergence.

\subsubsection{$f$-Divergence $h(\dvar)\defeq \fdiv(\dvar \| \visitrb)$}
\label{sec:fdiv-ope}
The use of an $f$-divergence objective leads to the set of general off-policy estimation techniques outlined in the recent \emph{DualDICE} paper~\citep{nachum2019dualdice}. Specifically, the various estimators proposed by DualDICE correspond to applying either the Lagrange or Fenchel-Rockafellar dualities to the optimization problem,
\begin{align}
    \label{eq:fdiv-ope-obj}
    \max_{\dvar} & ~-\fdiv(\dvar \| \visitrb) \\
    \label{eq:fdiv-const}
    \st &~ \dvar(s,a) = (1-\gamma)\init(s)\pi(a|s) + \bellmant \dvar(s,a), \\
    &~\forall s\in \Sset, a\in\Aset.
\end{align}

\paragraph{Lagrange Duality}
Application of Lagrange duality to the above problem yields 
\begin{multline}
    \max_\dvar \min_\qvar L(\qvar,\dvar) \\
    \defeq -\fdiv(\dvar \| \visitrb) + \sum_{s,a} \qvar(s,a)\cdot((1-\gamma)\init(s)\pi(a|s) +\bellmant\dvar(s,a) - \dvar(s,a)) \\
    =(1 - \gamma)\cdot \E_{\initsamp}[\qvar(s_0,a_0)] - \fdiv(\dvar \| \visitrb) + \sum_{s,a} \qvar(s,a)\cdot(\bellmant\dvar(s,a) - \dvar(s,a))
\end{multline}
We transform the transpose policy transition operator $\bellmantnog$ to its transpose $\bellmannog$ by using the fact $\langle y, A x\rangle = \langle x, A_* y\rangle$:
\begin{equation}
    L(\qvar,\dvar) =
    (1 - \gamma)\cdot \E_{\initsamp}[\qvar(s_0,a_0)] - \fdiv(\dvar \| \visitrb) + \sum_{s,a} \dvar(s,a)\cdot(\bellman\qvar(s,a) - \qvar(s,a))
\end{equation}
Now we make the change-of-variables $\zeta(s,a) = \frac{\dvar(s,a)}{\visitrb(s,a)}$ to yield,
\begin{multline*}
    \max_\zeta \min_\qvar L(\qvar,\zeta) \\
    \defeq (1 - \gamma)\cdot \E_{\initsamp}[\qvar(s_0,a_0)] - \E_{(s,a)\sim\visitrb}[f(\zeta(s,a))] + \E_{(s,a)\sim\visitrb}[\zeta(s,a)\cdot(\bellman\qvar(s,a) - \qvar(s,a))]
\end{multline*}
\begin{equation}
    \hspace{-1mm}
    \resizebox{.95\hsize}{!}{
    $= (1 - \gamma)\cdot \E_{\initsamp}[\qvar(s_0,a_0)] + \E_{(s,a,s')\sim\visitrb}[\zeta(s,a)\cdot(\bellman\qvar(s,a) - \qvar(s,a)) - f(\zeta(s,a))]$.
    }
    \hspace{-3mm}
    \label{eq:dualdice-zeta}
\end{equation}
Thus we have recovered the general saddle-point form of DualDICE, which proposes to optimize~\eqref{eq:dualdice-zeta} and then use an approximate solution $\hat{\zeta}^*$ to estimate $\avgstep(\pi)$ via~\eqref{eq:avgstep-zeta}.

\paragraph{Fenchel-Rockafellar Duality}
Rather than applying Lagrange duality, application of Fenchel-Rockafellar duality to~\eqref{eq:fdiv-ope-obj} more clearly reveals the wisdom of choosing $h(d)=\fdiv(\dvar \| \visitrb)$.
%We can also understood the~\eqref{eq:fdiv-ope-obj} as $f$-divergence regularized duality. 
We write the problem in~\eqref{eq:fdiv-ope-obj} as 
$$
\max_d\,\, -g(-A d)-h(d),
$$
where $g(-A d)$ corresponds to the linear constraints~\eqref{eq:fdiv-const} with respect to the adjoint Bellman operator; \ie,
$$g\defeq\indicator_{\{(1-\gamma)\init\times\pi\}} ~\text{and}~  A\defeq \bellmant - I.$$
When applying Fenchel-Rockafellar duality, the linear operator $A$ is transformed to its adjoint $A_* = \bellman - I$ and is used as an argument to the Fenchel conjugate $h_*(\cdot)=\E_{\visitrb}[\fstar(\cdot)]$ of $h$. At the same time, $g$ is replaced by its Fenchel conjugate $g_*(\cdot)=(1-\gamma)\E_{\init\times\pi}[\cdot]$.

The dual problem is therefore given by
\begin{align}
    &\min_{\qvar}~g_*(\qvar) + h_*(A_*\qvar) \\
    \label{eq:dualdice}
    =&\min_\qvar ~(1-\gamma)\cdot\E_{\initsamp}[\qvar(s_0,a_0)] +  \E_{(s,a)\sim\visitrb}[\fstar(\bellman\qvar(s,a) - \qvar(s,a))].
\end{align}
We can now see that the use of an $f$-divergence with respect to $\visitrb$ naturally leads to an offline problem with expectations over $\visitrb$, without an explicit change-of-variables.
Furthermore, unlike previous dual problems, there are no constraints in this optimization, and so standard gradient-based techniques may be applied to find a solution $\qvar^*$ without appealing to the Lagrange duality, which would necessarily involve nested $\max$-$\min$ optimizations.
The Fenchel-Rockafellar duality also provides us with a way to recover $\dvar^*=\visitpi$ from a solution $\qvar^*$:
\begin{equation}
    \visitrb(s,a) \cdot \fstar'(\bellman\qvar^*(s,a) - \qvar^*(s,a)) = \dvar^*(s,a),
\end{equation}
or equivalently,
\begin{equation}
     \fstar'(\bellman\qvar^*(s,a) - \qvar^*(s,a)) = \frac{\visitpi(s,a)}{\visitrb(s,a)}.
\end{equation}
If we set $f(x) = \frac{1}{2}x^2$, we may recover what is perhaps the most intriguing result in~\citet{nachum2019dualdice}:
\begin{multline}
    \qvar^* = \arg\min_{\qvar} ~(1-\gamma)\cdot\E_{\initsamp}[\qvar(s_0,a_0)] + \frac{1}{2}\E_{(s,a)\sim\visitrb}[(\bellman\qvar(s,a) - \qvar(s,a))^2] \\
    \Rightarrow \bellman\qvar^*(s,a) - \qvar^*(s,a) = \frac{\visitpi(s,a)}{\visitrb(s,a)}, ~~\forall s\in\Sset,a\in\Aset.
\end{multline}
That is, if one optimizes $Q$-value functions to minimize squared Bellman residuals (with respect to zero reward) while minimizing initial $Q$-values, then the optimal Bellman residuals are exactly the density ratios between the on-policy and offline state-action distributions.

Interestingly, the derivations in~\citet{nachum2019dualdice} do not explicitly use Lagrangian or Fenchel-Rockafellar duality, but rather focus on a cleverly chosen change-of-variables (the so-called \emph{DualDICE} trick). 
It is clear from our own derivations that this trick essentially comes from the relationship between $\bellmannog$ and $\bellmantnog$ and is simply another way of applying Fenchel-Rockafellar duality.

It is important to note that there is a trade-off introduced by the use of the Fenchel-Rockafellar duality as opposed to the Lagrangian. Namely, the objective~\eqref{eq:dualdice} involves optimizing a convex function of an expectation under $\bellmannog$, and thus also the environment transition function $T$, whereas in practice one typically only has access to a single empirical sample $s'\sim T(s,a)$. Many works ignore this problem, and simply consider the single empirical sample $s'$ as the full distribution $T(s,a)$, and this introduces a bias into the optimization. See~\citet{antos2008learning,dai2016learning} for potential remedies to this issue.

\subsection{Summary}
We briefly summarize the main takeaways from this section.
\begin{itemize}
    \item The policy evaluation problem may be expressed as an LP, known as the $Q$-LP, whose solution is $\qpi$.
    \item The dual of this LP has solution $\visitpi$. % , although it is over-constrained.
    \item Taking the Lagrangian of the $Q$-LP can lead to a doubly robust estimator for $\avgstep(\pi)$.
    \item Changing the objective in the dual of the $Q$-LP does not change its solution $\visitpi$. 
    \item Changing the objective to an appropriate alternative is a powerful tool. This can lead to a (regularized) Fenchel-Rockafellar dual that is unconstrained, and thus more amenable to stochastic and offline settings.
    \item These techniques may be used to derive the results provided in a number of recent works, such as~\citet{nachum2019dualdice} and~\citet{uehara2019minimax}.
\end{itemize}
\section{Policy Optimization}
\label{sec:opt}

% \Bo{TODO
% \begin{itemize}
%     \item cite SBEED and Tsallis-PCL
% \end{itemize}
% }

The previous section detailed a number of ways to estimate $\avgstep(\pi)$. In this section, we show how similar techniques may be applied for the policy optimization problem, which is concerned with finding the optimal solution $\arg\max_\pi \avgstep(\pi)$. 

\subsection{The Policy Gradient Theorem}
\label{sec:pg-theorem}
Considering the Lagrangian formulation of $\avgstep(\pi)$ given in~\eqref{eq:eval-lagrange} can provide a simple derivation of the policy gradient theorem in~\eqref{eq:policy-gradient}.
Let $L(\qvar,\dvar;\pi)$ be the inner expression in~\eqref{eq:eval-lagrange}.
Danskin's theorem~\citep{Bertsekas99} tells us that
\begin{equation}
    \frac{\partial}{\partial\pi}\avgstep(\pi) = \frac{\partial}{\partial\pi}\min_\qvar\max_{\dvar\ge0} L(\qvar,\dvar;\pi) = \frac{\partial}{\partial\pi} L(\qvar^*,\dvar^*;\pi),
\end{equation}
where $\qvar^*,\dvar^*$ are the solutions to $\min_\qvar\max_{\dvar\ge0} L(\qvar,\dvar;\pi) = \max_{\dvar\ge0}\min_\qvar L(\qvar,\dvar;\pi)$.
%Recall that $\qvar^*(s,a)=\qpi(s,a)$ for all $s,a$ for which $\visitpi(s,a)>0$ and that $\dvar^*=\visitpi$. Thus,
%\begin{equation}
%    \frac{\partial}{\partial\pi}\avgstep(\pi) = \frac{\partial}{\partial\pi} L(\qpi,\visitpi;\pi).
%\end{equation}
We may compute the gradient of $L(\qvar^*,\dvar^*;\pi)$ w.r.t. $\pi$ term-by-term. For the first term in~\eqref{eq:eval-lagrange} we have
\begin{equation}
    \label{eq:pg-first-term}
    \frac{\partial}{\partial\pi}(1-\gamma)\cdot\E_{\initsamp}[\qvar^*(s_0,a_0)] = (1-\gamma)\cdot\E_{\initsamp}[\qvar^*(s_0,a_0)\nabla\log\pi(a_0|s_0)],
\end{equation}
where we used the general identity $\frac{\partial}{\partial p} \E_{z\sim p}[h(z)] = \E_{z\sim p}[h(z)\nabla\log p(z)]$. For the second term of $L$ in~\eqref{eq:eval-lagrange} we have
\begin{equation*}
    \frac{\partial}{\partial\pi}\E_{(s,a)\sim\dvar^*}[\rew(s,a) + \bellman\qvar^*(s,a) - \qvar^*(s,a)] = \E_{(s,a)\sim\dvar^*}\left[\gamma\cdot\frac{\partial}{\partial\pi}\E_{\substack{s'\sim T(s,a) \\ a'\sim\pi(s')}}[\qvar^*(s',a')]\right]
\end{equation*}
\begin{equation}
    \label{eq:pg-second-term}
    = \gamma\cdot\E_{\substack{(s,a)\sim\dvar^*,s'\sim T(s,a) \\ a'\sim\pi(s')}}[\qvar^*(s',a')\nabla\log\pi(a'|s')].
\end{equation}
Recall that $\qvar^*(s,a)=\qpi(s,a)$ for all $s,a$ for which $\visitpi(s,a)>0$ and that $\dvar^*=\visitpi$. Furthermore,
\begin{equation}
    \visitpi(s,a) = (1-\gamma)\init(s)\pi(a|s) +\gamma \pi(a|s)\sum_{\tilde{s},\tilde{a}}T(s'|\tilde{s},\tilde{a})\visitpi(\tilde{s},\tilde{a}).
\end{equation}
Using this information, we may combine~\eqref{eq:pg-first-term} and~\eqref{eq:pg-second-term} to yield
\begin{equation}
    \frac{\partial}{\partial\pi}L(\qvar^*,\dvar^*;\pi) = \E_{(s,a)\sim\visitpi}[\qpi(s,a)\nabla\log\pi(a|s)],
\end{equation}
which matches the policy gradient theorem as presented in~\eqref{eq:policy-gradient}.

\subsection{Offline Policy Gradient via the Lagrangian}
The original policy gradient theorem in~\citet{sutton2000policy} relies on on-policy settings, while in practice we would like to optimize the policy with only \emph{offline} samples.
As in Section~\ref{sec:eval-lagrange}, one may write $\avgstep(\pi)$ in an offline manner via the variables $\zeta(s,a)=\frac{\visitpi(s,a)}{\visitrb(s,a)}$.
Using~\eqref{eq:ope-lagrange}, we may write the policy optimization problem as
\begin{multline}
    \max_\pi\min_\qvar\max_{\zeta\ge0} L(Q,\zeta,\pi) \\
    \hspace{-2mm}\defeq (1-\gamma)\cdot\E_{\initsamp}[\qvar(s_0,a_0)] +  \E_{\substack{(s,a,s')\sim\visitrb \\ a'\sim\pi(s')}}[\zeta(s,a)\cdot(\rew(s,a) + \gamma\qvar(s',a') - \qvar(s,a))].\hspace{-2mm}
\end{multline}
As in Section~\ref{sec:pg-theorem}, one may argue that if $\qvar,\zeta$ are optimized, then the gradient $\frac{\partial}{\partial\pi} L(Q,\zeta,\pi)$ will be exactly the on-policy policy gradient $\E_{(s,a)\sim\visitpi}[\qpi(s,a)\nabla\log\pi(a|s)]$, thus achieving on-policy policy gradient with offline data.

\subsection{Fenchel-Rockafellar Duality for Regularized Optimization}
%As discussed in Section~\ref{sec:change-the-problem}, optimization via the Lagrangian is hard to avoid when both the primal and the dual involve constraints which are difficult to handle in stochastic optimization settings.
% Application of the Lagrangian is necessary when both the primal and dual involve constraints, especially in the RL setting where the constraints are numerous and thus difficult to handle in stochastic optimization settings.
As noted in previous sections, both the linear nature and $\min$-$\max$ form of the Lagrangian may lead to numerical instability in practice. In Section~\ref{sec:fdiv-ope}, we showed how regularizing the objective of the LP for $\visitpi$ in an appropriate manner can lead to a Fenchel-Rockafellar dual that is unconstrained (and thus avoids the need for the Lagrangian).
In this previous section, we only cared about estimating $\dvar^*=\visitpi$, and conveniently, the regularization did not affect the optimal solution.
However, in our current setting of policy optimization, changing the objective will change the optimal policy.
Still, the modified objective may be motivated as a regularization of the max-reward policy objective, and finding the optimal regularized policy is still desirable in many applications.

Thus, we consider regularizing the max-reward policy objective with the $f$-divergence $\fdiv(\dvar \| \visitrb)$. Our modified problem becomes
\begin{align}
    \label{eq:fdiv-obj}
    \avgstep(\pi) - \fdiv(\visitpi\|\visitrb) = \max_{\dvar} & ~ - \fdiv(\dvar \| \visitrb) + \sum_{s,a}\dvar(s,a)\cdot \rew(s,a) \\
    \st &~ \dvar(s,a) = (1-\gamma)\init(s)\pi(a|s) + \bellmant \dvar(s,a), \\
    &~\forall s\in \Sset, a\in\Aset.
\end{align}
Fenchel-Rockafellar duality yields the following dual formulation:
\begin{multline}
    \avgstep(\pi) -\fdiv(\visitpi\|\visitrb) = \min_\qvar ~(1-\gamma)\cdot\E_{\initsamp}[\qvar(s_0,a_0)] + \\ \E_{(s,a)\sim\visitrb}[\fstar(\rew(s,a) + \bellman\qvar(s,a) - \qvar(s,a)) ].
\end{multline}
Thus, one may optimize $\pi$ to maximize the regularized objective by solving the $\max$-$\min$ optimization
\begin{equation}
    \label{eq:algae}
    \max_\pi\min_\qvar ~(1-\gamma)\cdot\E_{\initsamp}[\qvar(s_0,a_0)] + \E_{(s,a)\sim\visitrb}[\fstar(\rew(s,a) + \bellman\qvar(s,a) - \qvar(s,a)) ],
\end{equation}
and the optimal solution $\pistar$ of this objective is the optimal solution to the regularized problem $\max_\pi \avgstep(\pi) -\fdiv(\visitpi \| \visitrb)$.
This recovers the offline formulation of regularized policy gradient known as~\emph{Primal AlgaeDICE} derived in~\citet{algae}. 
As noted in~\citet{algae}, when using $f(x)=\frac{1}{2}x^2$, the objective bears some resemblance to actor-critic learning but with offline samples, in which a $Q$-value is learned to minimize squared Bellman errors and a policy $\pi$ is learned to maximize $Q$-values.

Although the optimal $\pistar$ of the objective in~\eqref{eq:algae} is affected by the regularization $\fdiv(\dvar\|\visitrb)$, the optimal $\dvar^*=\visitpi$ is not (since the problem in~\eqref{eq:fdiv-obj} is still over-constrained). Thus, one can show that the objective in~\eqref{eq:algae} still possesses the on-policy policy gradient with offline data property; \ie, if $\qvar$ is optimized, the gradient of the objective with respect to $\pi$ is the on-policy policy gradient, although the policy gradient is with respect to regularized (modified) $\tilde{Q}$-values whose exact values depend on the form of $f$.

Of course, one may also write the regularized problem~\eqref{eq:fdiv-obj} in its Lagrangian form, and this can recover \emph{Fenchel AlgaeDICE} as derived in~\citet{algae}.

\subsubsection{Regularization with the $\KL$-Divergence}
\label{sec:reg-kl}
If one considers regularizing the max-reward objective with the $\KL$-divergence $\dkl(\dvar\|\visitrb)$, one arrives at the constrained optimization problem,
\begin{align}
    \label{eq:kl-obj}
    \avgstep(\pi) - \dkl(\visitpi\|\visitrb) = \max_{\dvar\in\Delta(|\Sset|\times|\Aset|)} & ~ - \dkl(\dvar \| \visitrb) + \sum_{s,a}\dvar(s,a)\cdot \rew(s,a) \\
    \st &~ \dvar(s,a) = (1-\gamma)\init(s)\pi(a|s) + \bellmant \dvar(s,a), \\
    &~\forall s\in \Sset, a\in\Aset.
\end{align}
Subsequent application of Fenchel-Rockafellar duality leads to the following offline policy optimization objective:
\begin{equation*}
    \max_\pi\min_\qvar (1-\gamma)\cdot\E_{\initsamp}[\qvar(s_0,a_0)] + \log \E_{(s,a)\sim\visitrb}[\exp\{\rew(s,a) + \bellman\qvar(s,a) - \qvar(s,a)\} ].
\end{equation*}

The beauty of this objective is revealed when one considers optimizing $\pi$ with respect to a gradient-based method. For a specific $\qvar$, the gradient of this objective with respect to $\pi$ is
\begin{multline}
    (1-\gamma)\cdot\E_{\initsamp}[\qvar(s_0,a_0)\nabla\log\pi(a_0|s_0)] \\ + \gamma\cdot\E_{\substack{(s,a,s')\sim\visitrb \\ a'\sim\pi(s')}}[\softmax_{\visitrb}(\rew + \bellman\qvar - \qvar)(s,a)\cdot \qvar(s',a')\nabla\log\pi(a'|s')],
\end{multline}
where the softmax function is defined as
\begin{equation}
    \softmax_{p}(h)(z) \defeq \frac{\exp\{h(z)\}}{\E_{\tilde{z}\sim p}[\exp\{h(\tilde{z}\}]}.
\end{equation}

Thus, we see that the use of KL-divergence leads to a dual formulation which bears similarities to max-likelihood policy learning, a common goal of a number of recent works~\citep{abdolmaleki2018maximum,song2019v,peng2019advantageweighted}.
The use of max-likelihood policy learning in conjunction with a log-average-exp objective for the $Q$-value function also bears strong resemblance to the REPS algorithm~\citep{peters2010relative}.
Still, the policy objective here only resembles max-likelihood learning and is not exactly equivalent. The connections hinted at here will be made more explicit in~\secref{sec:vlp}.

\subsection{Imitation Learning}

The $f$-divergence regularized dual form of the $Q$-LP in~\eqref{eq:fdiv-obj} can also be applied for imitation learning, in which one considers the offline dataset $\Dset$ to be demonstrations from an expert policy, and the goal is to recover this expert policy. 
Indeed, if one ignores rewards (or considers them to be zero) in~\eqref{eq:fdiv-obj}, the optimization corresponds to finding a policy $\pi$ which minimizes the $f$-divergence in terms of the state-action occupancies from $\visitrb$, \ie, $D_f\rbr{\visitpi\|\visitrb}$.
Many existing imitation learning works~\citep{ho2016generative,fu2017learning,kostrikov2018discriminator,ke2019imitation,ghasemipour2019divergence} aim to exactly optimize this $f$-divergence objective. However, these methods typically rely on online access to the environment. With the same techniques as we applied for offline policy evaluation and offline policy optimization, one can derive offline imitation learning algorithms. The derived objectives will be similar to Primal AlgaeDICE or Fenchel AlgaeDICE without rewards $R(s,a)$. Particularly, \citet{kostrikov2019imitation} instantiates the offline imitation learning following such a strategy with the $\KL$-divergence.

\subsection{Summary}
We briefly summarize the main takeaways from this section.
\begin{itemize}
    \item One can apply many of the same techniques used for policy evaluation to the policy optimization problem by simply putting a $\max_\pi$ around a derived estimator for $\avgstep(\pi)$.
    \item Since the optimal solution of the inner optimization will typically be either $\visitpi$ or $\visitpi/\visitrb$, one can appeal to Danskin's theorem to argue that the gradients of $\pi$ will be on-policy policy gradients, even if one only has access to offline data.
    \item We again see the power of modifying an objective before appealing to duality. Appropriate regularization of the max-reward objective can lead to an unconstrained Fenchel-Rockafellar dual problem.
    \item The same technique can also be exploited for offline imitation learning, which can be simply derived from the policy optimization objectives by ignoring the reward.
    \item Depending on the exact form of regularization, application of Fenchel-Rockafellar duality leads to objectives which hint at connections to actor-critic via squared Bellman error minimization as well as max-likelihood policy learning. More interesting or useful objectives may be possible for other (yet undiscovered?) specially chosen regularizers.
\end{itemize}

\section{RL with the Linear Programming Form of $V$}
\label{sec:vlp}

The policy optimization approaches of the previous section may seem superficial. We essentially took formulations of $\avgstep(\pi)$ from Section~\ref{sec:eval} and put a $\max_\pi$ around them.
While this is valid, it leads to problems-within-problems, \ie $\max$-$\min$ problems, for which stochastic optimizations can be difficult to theoretically motivate. Is there a better way?

There is a more direct way to frame the policy optimization problem as a convex problem, and it is given by the $V$-LP~\citep{puterman1994markov,bertsekas1995dynamic,bertsekas1996neuro,wang2007dual}. We begin by introducing what is usually referred to as the dual of the $V$-LP, expressed in terms of state-action visitations:
\begin{align}
    \label{eq:vlp-dual}
    \max_{\dvar\ge0} & ~\sum_{s,a} \dvar(s,a)\cdot\rew(s,a) \\
    \label{eq:vlp-dual-const}
    \st &~ \sum_a \dvar(s,a) = (1-\gamma)\init(s) + \vbellmant \dvar(s), \\
    &~\forall s\in \Sset,
\end{align}
where we use $\vbellmantnog$ to denote the transpose (or adjoint) transition operator,
\begin{equation}
    \label{eq:vbellmant}
    \vbellmantnog\dvar(s) \defeq \sum_{\tilde{s},\tilde{a}} T(s|\tilde{s},\tilde{a})\cdot \dvar(\tilde{s},\tilde{a}). 
\end{equation}
As in~\eqref{eq:d-const}, the constraints in~\eqref{eq:vlp-dual-const} describe the conservation of flow (probability mass) of a stationary distribution on a Markovian process, although now the conservation is measured with respect to states as opposed to state-action pairs.
Crucially, unlike the problem in~\eqref{eq:d-obj}, this problem is not over-constrained. The result of this problem is $\avgstep(\pistar)$ for an optimal max-reward policy $\pistar$ and the solution is $\dvar^* = \visitpistar$.

The dual of problem~\eqref{eq:vlp-dual} is the more commonly seen $V$-LP (which is typically referred to as the primal):
\begin{align}
    \label{eq:vlp-primal}
    \min_{\vvar} & ~(1-\gamma)\cdot\E_{s_0\sim\init}[\vvar(s_0)] \\
    \label{eq:vlp-primal-const}
    \st &~ \vvar(s) \ge \rew(s,a) + \vbellman \vvar(s,a), \\
    &~\forall s\in \Sset, a\in\Aset,
\end{align}
where now $\vbellmannog$ is the transition operator,
\begin{equation}
    \label{eq:vbellman}
    \vbellmannog\vvar(s,a) \defeq \E_{s'\sim T(s,a)}[\vvar(s')]. 
\end{equation}
The optimal $\vvar^*$ of this problem is the value function of an optimal policy $V^{\pistar}$, where
\begin{equation}
    V^\pi(s) \defeq \E_{a\sim\pi(s)}[\qpi(s,a)].
\end{equation}

With these primal and dual representations of the policy optimization problem, we may apply the same techniques of the previous sections.\footnote{The Lagrangian of~\eqref{eq:vlp-primal} or~\eqref{eq:vlp-dual} may also be used to derive several existing~\emph{tabular}~\citep{chen2016stochastic} and \emph{on-policy} algorithms~\citep{wang2017randomized,dai2017boosting,chen2018scalable}.}
%, leading to several algorithms in \citet{chen2016stochastic,wang2017randomized,dai2017boosting}.
However, different from the $Q$-LP, since the problem is not over-constrained by the $|\Sset|$ constraints in~\eqref{eq:vlp-dual-const}, one cannot ignore the additional $|\Sset|\times|\Aset|$ constraints $d\ge 0$ in~\eqref{eq:vlp-dual}.
For example, if one augments the problem in~\eqref{eq:vlp-dual} with a general regularizer $-\fdiv(\dvar\|\visitrb)$, one will have to consider $d\ge0$ as an additional set of linear constraints. This will lead to a dual objective over two functions: $\vvar:\Sset\to\R$ and $K:\Sset\times\Aset\to\R_+$:
\begin{equation}
    \label{eq:fdiv-v-obj}
    \min_{K\ge0,\vvar} ~(1-\gamma)\cdot\E_{s_0\sim\init}[\vvar(s_0)] + \E_{(s,a)\sim\visitrb}[\fstar(K(s,a) + \rew(s,a) + \vbellman\vvar(s,a) - \vvar(s))].
\end{equation}
This objective can be seen as an improvement over our previous derivation using the $Q$-LP, since, unlike before, this objective only involves a single optimization over $\vvar,K$ as opposed to a $\max$-$\min$ optimization over $\pi$ and $\qvar$.
However, the solution to this problem will give us $\vvar^*$, the (regularized) value function of the optimal (regularized) policy, while what we really want is the policy itself!

To derive the optimal policy, we first note that Fenchel-Rockafellar duality provides us the optimal $\dvar^*$ from $\vvar^*,K^*$ via
\begin{equation}
    \dvar^*(s,a) = \visitrb(s,a)\cdot\fstar'(K^*(s,a) + \rew(s,a) + \vbellman\vvar^*(s,a) - \vvar^*(s)).
\end{equation}
The solution $\dvar^*$ is the visitation of an optimal regularized policy $\visitpistar$. Using Bayes' rule we may find the optimal policy to be
\begin{equation*}
    \pistar(a|s) = \frac{\dvar^*(s,a)}{\sum_{\tilde{a}} \dvar^*(s,\tilde{a})}
    = \frac{\visitrb(s,a)\cdot \fstar'(K^*(s,a) + \rew(s,a) + \vbellman\vvar^*(s,a) - \vvar^*(s))}{\sum_{\tilde{a}} \visitrb(s,\tilde{a})\cdot\fstar'(K^*(s,a) + \rew(s,\tilde{a}) + \vbellman\vvar^*(s,\tilde{a}) - \vvar^*(s))}.
\end{equation*}
This way, one may recover the optimal policy $\pistar$ from the solutions $\vvar^*,K^*$ of~\eqref{eq:fdiv-v-obj}. This form of value learning (and $K$ learning) before recovering a policy from the optimal values is the same as derived by~\citet{belousov2017f}.

We can see that the use of the Fenchel-Rockafellar dual of $V$-LP introduces a trade-off compared to starting from the $Q$-LP. We have avoided nested $\max$-$\min$ optimizations that arise from the $Q$-LP, but now our problems do not directly provide us with a policy. One must perform extra derivations to derive the policy. Depending on the specific regularization employed, deriving the optimal policy from $\vvar^*,K^*$ may be difficult in practical (stochastic) settings. 
%While the Lagrangian of $V$-LP~\eqref{eq:vlp-primal} will introduce explicit optimization to the policy, avoiding the extra policy recovery, comparing to the algorithm derived from the $Q$-LP, the algorithms from $V$-LP is not able to handle the \emph{behavior-agnostic} offline RL setting. 

\subsection{Max-Likelihood Policy Learning}
Regularization via the $\KL$-divergence provides for a practical way to avoid the numerical instability and ensure the positiveness of $\dvar$. %~\citep{neu2017unified}.
% and intriguing way to recover the optimal policy. 
This was hinted at previously in Section~\ref{sec:reg-kl} with respect to the $Q$-LP.

For general $f$-divergence, the constraints $d\ge0$, must be considered as additional linear constraints. However, the dual form of the $\KL$ as a log-expected-exponent is already with respect to a non-negative domain (more specifically, $d\in\Delta(\Sset\times\Aset)$). Therefore, regularization of~\eqref{eq:vlp-dual} with $\dkl(\dvar\|\visitrb)$ and application of Fenchel-Rockafellar duality provides us with the simple objective
\begin{equation}
    \min_{\vvar}~(1-\gamma)\cdot\E_{s_0\sim\init}[\vvar(s_0)] + \log\E_{(s,a)\sim\visitrb}[\exp\{\rew(s,a) + \vbellman\vvar(s,a) - \vvar(s)\}],
\end{equation}
and this recovers the REPS objective~\citep{peters2010relative}.\footnote{See also~\citet{neu2017unified}.}

The visitations of the optimal policy are now given by the $\softmax$ function:
\begin{equation}
    \visitpistar(s,a) = \visitrb(s,a)\cdot\softmax_{\visitrb}(\rew + \vbellman\vvar^* - \vvar^*)(s,a).
\end{equation}
The optimal policy thus has a similar form:
\begin{equation}
    \pistar(a|s) = \visitrb(a|s)\cdot\softmax_{\visitrb(\cdot|s)}(\rew(s,\cdot) + \vbellman\vvar^*(s,\cdot) - \vvar^*(s))(a).
\end{equation}
We may now see that recovering $\pistar$ from $\vvar^*$ may be done via max-likelihood learning:
\begin{equation}
    \pistar = \arg\max_\pi \E_{(s,a)\sim\visitrb}\left[\frac{1}{Z(s)}\exp\{\rew(s,a)+\vbellman\vvar^*(s,a) - \vvar^*(s)\}\log\pi(a|s)\right],
\end{equation}
where $Z(s)$ is an arbitrary normalization.
This either recovers or can be used to motivate many past and recent works advocating for max-likelihood policy learning~\citep{peters2010relative,abdolmaleki2018maximum,song2019v,peng2019advantageweighted}

\subsection{Policy Evaluation with the $V$-LP}
Although we originally introduced the $V$-LP as a way to perform more streamlined policy optimization, one can also use it for policy evaluation. To do so, we decompose $\dvar(s, a) = \mu(s)\pi(a|s)$ for a fixed policy $\pi(a|s)$. Plugging this form of $\dvar(s, a)$ into~\eqref{eq:vlp-dual} leads to 
\begin{align}
    \label{eq:vlp-dual-decomposed}
    \max_{\mu} & ~\sum_{s,a} \mu(s)\pi(a|s)\cdot\rew(s,a) \\
    \label{eq:vlp-dual-const-decomposed}
    \st &~ \mu(s) = (1-\gamma)\init(s) + \vbellmant (\mu\times\pi)(s), \\
    &~\forall s\in \Sset.
\end{align}
When one fixes the policy $\pi(a|s)$ in this manner, the resulting LP is over-constrained, in the sense that the $\abr{\Sset}$ constraints in~\eqref{eq:vlp-dual-const-decomposed} uniquely determine the on-policy state visitation $\mu^\pi(s)$. Applying the Lagrangian to~\eqref{eq:vlp-dual-decomposed} will result in objectives similar to those that have appeared in~\citet{kallus2019efficiently,tang2019doubly}. 
Moreover, as we have done in~\secref{sec:change-the-problem}, one can also replace the objective~\eqref{eq:vlp-dual-decomposed} before applying either Lagrangian or Fenchel-Rockafellar duality. 
One should note that the use of a $V$-LP for offline policy evaluation results in objectives that require knowledge of the data distribution policy $\visitrb(a|s)$ in order to correct the offline action samples to on-policy samples, as needed by the use of $\mu\times\pi$ in~\eqref{eq:vlp-dual-const-decomposed}. This is in contrast to the \emph{behavior-agnostic} objectives yielded by the $Q$-LP.
%emphasize that although the difference between~\eqref{eq:vlp-dual-decomposed} and~\eqref{eq:d-obj} is slight, it is crucial to enable the \emph{behavior-agnostic} policy evaluation. 

\section{Undiscounted Settings}
\label{sec:undiscounted}
So far, our treatment has only considered discounted settings, \ie, $\gamma\in(0, 1)$.  The undiscounted setting $\gamma=1$ presents an interesting challenge for many RL algorithms, as notions of $Q$-values and convergence of Bellman backup operators are harder to get a handle on.
On the other hand, the techniques of Fenchel-Rockafeller duality and its variants may be readily applied to the undiscounted setting, with only minor modifications. In this section, we provide an extension of our previous derivations to the $\gamma=1$ case, leading to several practical algorithms.

\subsection{Policy Evaluation}
When $\gamma=1$, the policy evaluation problem concerns estimating the average per-step reward of the policy:
\begin{equation}
    \avgstep(\pi) \defeq \lim_{t_{\mathrm{stop}}\to\infty} \E\left[\frac{1}{t_{\mathrm{stop}}}\left.\sum_{t=0}^{t_{\mathrm{stop}}} \rew(s_t,a_t) ~\right| s_0\sim\init, \forall t, a_t\sim \pi(s_t), s_{t+1}\sim T(s_t, a_t)\right].
\end{equation}
Under certain conditions,\footnote{In finite state and action spaces, the MDP must be ergodic. For other (continuous) spaces, the conditions are more involved. Please refer to~\citet{zhang2020gendice} for the details.} this quantity may be alternatively written as an expectation over the stationary distribution of $\pi$ in $\mdp$:
\begin{equation}\label{eq:undisc-rho}
    \avgstep(\pi) = \E_{(s,a)\sim\visitpi}[\rew(s,a)],
\end{equation}
where the undiscounted on-policy distribution $\visitpi$ is defined as the normalized distribution satisfying
% It is the unique (normalized) distribution satisfying
\begin{equation}
    \visitpi(s,a) = \bellmantnog\visitpi(s,a) ~~\forall s\in\Sset,a\in\Aset.
\end{equation}
Thus, we can formulate $\rho(\pi)$ analogous to~\eqref{eq:d-obj} as,
\begin{align}
    \label{eq:undisc-d-obj}
    \avgstep(\pi) = \max_{\dvar\ge0} & ~\sum_{s,a} \dvar(s,a)\cdot\rew(s,a) \\
    \label{eq:undisc-d-const}
    \st &~ \sum_{s,a}\dvar(s,a) = 1
    ~\text{and}~\\
    &~\dvar(s,a) = \bellmantnog \dvar(s,a), 
    ~\forall s\in \Sset, a\in\Aset.
\end{align}
Note that the only difference to~\eqref{eq:d-obj} is that this LP requires the constraint~\eqref{eq:undisc-d-const}, which ensures that $\dvar$ constitutes a normalized distribution. Still, as in~\eqref{eq:d-obj}, this problem is over-constrained. The objective in~\eqref{eq:undisc-d-obj} may be modified without changing the solution $\dvar^*=\visitpi$, and so the same techniques used in Section~\ref{sec:eval} may be applied.

For example, if we modify the objective to $\max_{\dvar} -\fdiv(\dvar\|\visitrb)$ and then take the Fenchel-Rockafellar dual we arrive at the problem
\begin{equation}
    \label{eq:undisc-fdiv}
    \min_{\qvar,\lambda} ~-\lambda + \E_{(s,a)\sim\visitrb}[\fstar(\lambda + \bellmannog\qvar(s,a) - \qvar(s,a))].
\end{equation}
Given a solution $\qvar^*,\lambda^*$ to this problem, the optimal $\dvar^*=\visitpi$ may be recovered as
\begin{equation}
    \visitrb(s,a)\cdot\fstar'(\lambda^*+\bellmannog\qvar^*(s,a)-\qvar^*(s,a)) = \dvar^*(s,a),
\end{equation}
or equivalently,
\begin{equation}
    \fstar'(\lambda^*+\bellmannog\qvar^*(s,a)-\qvar^*(s,a)) = \frac{\visitpi(s,a)}{\visitrb(s,a)}.
\end{equation}

If one were to instead use the Lagrange duality with the objective $\max_{\dvar} -\fdiv(\dvar\|\visitrb)$, after making the change-of-variables $\zeta(s,a)=\frac{\dvar(s,a)}{\visitrb(s,a)}$ one would arrive at the nested optimization
\begin{equation}
    \label{eq:undisc-lagrange}
    \max_{\zeta}\min_{\qvar,\lambda} ~-\lambda + \E_{\substack{(s,a,s')\sim\visitrb \\ a'\sim\pi(s')}}[\zeta(s,a)\cdot(\lambda + \qvar(s',a') - \qvar(s,a)) - f(\zeta(s,a))].
\end{equation}
The optimal solution of this optimization is
\begin{equation}\label{eq:opt-dual}
    \zetastar(s,a) = \frac{\visitpi(s,a)}{\visitrb(s,a)}.
\end{equation}

\subsubsection{Regularized Lagrangian}
% According to the dual LP for the undiscounted policy value~\eqref{eq:undisc-d-obj}, we can also derive the primal form as
% \begin{align}
%     \label{eq:undisc-q-lp}
%     \avgstep(\pi) = \min_{\qvar, \lambda} & ~  \lambda \\
%     \st &~Q(s,a) \ge -\lambda + R\rbr{s, a}  + \bellman Q(s,a), 
%     ~\forall s\in \Sset, a\in\Aset.
% \end{align}
We refer the reader to~\citet{zhang2020gendice} for an alternative approach to undiscounted policy evaluation.
This approach essentially considers regularizing $\qvar$ and $\lambda$ in the Lagrangian of~\eqref{eq:undisc-d-obj} with a square regularization, eventually yielding the objective, 
\begin{equation}
    \label{eq:reg-undisc-lagrange}
    \max_{\zeta}\min_{\qvar,\lambda} ~-\lambda + \frac{1}{2}\lambda^2 + \E_{\substack{(s,a,s')\sim\visitrb \\ a'\sim\pi(s')}}[\zeta(s,a)\cdot(\lambda + \qvar(s',a') - \qvar(s,a) + \frac{1}{4}\qvar^2(s,a))) ].
\end{equation}
This corresponds to the objective used in~\emph{GenDICE}\footnote{In GenDICE, the objective is actually for $(\tilde\qvar, \tilde\lambda)\defeq (-\qvar, -\lambda)$ in~\eqref{eq:reg-undisc-lagrange}.} proposed in~\citet{zhang2020gendice} with $\chi^2$-divergence and the weight of the normalizing regularization set to be $1$. As discussed in~\citet{zhang2020gendice}, the regularizations on $\qvar, \lambda$ do not affect the optimal solution to the dual variable, \ie,~\eqref{eq:opt-dual} still holds. Therefore, we can still estimate the value of a policy via~\eqref{eq:undisc-rho}.
% \begin{equation}
% \zeta^*(s, a) = \frac{\visitpi(s,a)}{\visitrb(s,a)}.
% \end{equation}

\subsection{Policy Optimization}
As in Section~\ref{sec:opt}, we may approach the policy optimization problem by simply putting a $\max_\pi$ around a reward-aware form of~\eqref{eq:undisc-fdiv} or~\eqref{eq:undisc-lagrange}; \eg,
\begin{equation}
    \label{eq:po-undisc-fdiv}
    \max_\pi\min_{\qvar,\lambda} ~-\lambda + \E_{(s,a)\sim\visitrb}[\fstar(\lambda + \rew(s,a) + \bellmannog\qvar(s,a) - \qvar(s,a))].
\end{equation}
Similar to Section~\ref{sec:opt}, one may use Danskin's theorem to argue that the gradient with respect to $\pi$ of this objective is the on-policy policy gradient (although with respect to regularized $Q$-values).

Alternatively, one may use the techniques in Section~\ref{sec:vlp}, writing the policy optimization problem analogous to~\eqref{eq:vlp-dual} as
\begin{align}
    \label{eq:undisc-vlp-dual}
    \max_{\dvar\ge0} & ~\sum_{s,a} \dvar(s,a)\cdot\rew(s,a) \\
    \label{eq:undisc-vlp-dual-const}
    \st &~ \sum_{s,a}\dvar(s,a) = 1
    ~\text{and}~\\
    &~\sum_a \dvar(s,a) = \vbellmantnog \dvar(s),
    ~\forall s\in \Sset.
\end{align}
One may solve this problem via its Lagrangian or by adding an appropriate regularizer and applying Fenchel-Rockafellar duality~\citep{neu2017unified}.
For example, if one chooses to add a regularizer $-\dkl(\dvar\|\visitrb)$ to the objective in~\eqref{eq:undisc-vlp-dual}, then the Fenchel-Rockafellar dual is given by
\begin{equation}
\min_{\vvar}~\log\E_{(s,a)\sim\visitrb}[\exp\{\rew(s,a) + \vbellmannog\vvar(s,a) - \vvar(s)\}].
\end{equation}
The optimal policy $\pistar$ to the original problem may then be recovered from $\vvar^*$ via the max-likelihood optimization
\begin{equation}
    \pistar = \arg\max_\pi ~\E_{(s,a)\sim\visitrb}\left[\frac{1}{Z(s)}\exp\{\rew(s,a)+\vbellmannog\vvar^*(s,a) - \vvar^*(s)\}\log\pi(a|s)\right].
\end{equation}
\section{Conclusion}\label{sec:conclusion}
We have presented a variety of ways to apply the Fenchel-Rockafellar duality to problems appearing in RL. Although our settings and corresponding results are numerous, the techniques we used can be summarized succinctly as,
\begin{itemize}
    \item when presented with a problem that appears difficult to solve, consider writing the problem as a constrained convex optimization and solving its Fenchel-Rockafellar dual, or its Lagrangian form;
    \item if the dual is still difficult to solve (\eg, when the primal objective is linear, yielding a dual with constraints), consider modifying the original objective, \eg, by applying an appropriate convex regularizer.
\end{itemize}
This simple protocol has been a recurring theme in our derivations, leading to several algorithms to tackle the policy evaluation, policy optimization, and imitation learning problems regardless of online or offline access to the environment and discounted or undiscounted rewards. 

We hope the connections exposed here between RL and optimization can ignite progress and collaborations of researchers from both communities. For example, applying the same protocol outlined here to problems in other RL settings (\eg, multi-agent RL, safe RL, exploration for RL, \etc) and other ways to appropriately regularize these problems are promising RL research directions. 
At the same time, questions of how well these new formulations interact with algorithms for convex optimization (especially in stochastic and function approximation settings), and whether these duality-based formulations are more efficient than DP-based approaches, bring new challenges and problems for optimization research.

\subsubsection*{Acknowledgments}
We thank Lihong Li, Dale Schuurmans, Ilya Kostrikov, Yinlam Chow, Sherry Yang, George Tucker, Matthieu Geist, and other members of the Google Brain team for insightful thoughts and discussions.

\bibliographystyle{plainnat}
\bibliography{./ref}

\begin{thebibliography}{62}
\providecommand{\natexlab}[1]{#1}
\providecommand{\url}[1]{\texttt{#1}}
\expandafter\ifx\csname urlstyle\endcsname\relax
  \providecommand{\doi}[1]{doi: #1}\else
  \providecommand{\doi}{doi: \begingroup \urlstyle{rm}\Url}\fi

\bibitem[Abdolmaleki et~al.(2018)Abdolmaleki, Springenberg, Tassa, Munos,
  Heess, and Riedmiller]{abdolmaleki2018maximum}
Abbas Abdolmaleki, Jost~Tobias Springenberg, Yuval Tassa, Remi Munos, Nicolas
  Heess, and Martin Riedmiller.
\newblock Maximum a posteriori policy optimisation.
\newblock \emph{arXiv preprint arXiv:1806.06920}, 2018.

\bibitem[Ali and Silvey(1966)]{AliSil66}
S.M. Ali and S.D. Silvey.
\newblock A general class of coefficients of divergence of one distribution
  from another.
\newblock \emph{Journal of the Royal Statistical Society. Series B
  (Methodological)}, 28\penalty0 (1):\penalty0 131--142, 1966.

\bibitem[Antos et~al.(2008)Antos, Szepesv{\'a}ri, and Munos]{antos2008learning}
Andr{\'a}s Antos, Csaba Szepesv{\'a}ri, and R{\'e}mi Munos.
\newblock Learning near-optimal policies with bellman-residual minimization
  based fitted policy iteration and a single sample path.
\newblock \emph{Machine Learning}, 71\penalty0 (1):\penalty0 89--129, 2008.

\bibitem[Bas-Serrano and Neu(2019)]{bas2019faster}
Joan Bas-Serrano and Gergely Neu.
\newblock Faster saddle-point optimization for solving large-scale markov
  decision processes.
\newblock \emph{arXiv preprint arXiv:1909.10904}, 2019.

\bibitem[Bauschke and Lucet(2012)]{bauschke2012fenchel}
H~Bauschke and Yves Lucet.
\newblock What is a fenchel conjugate.
\newblock \emph{Notices of the AMS}, 59\penalty0 (1):\penalty0 44--46, 2012.

\bibitem[Bellman(1966)]{bellman1966dynamic}
Richard Bellman.
\newblock Dynamic programming.
\newblock \emph{Science}, 153\penalty0 (3731):\penalty0 34--37, 1966.

\bibitem[Belousov and Peters(2017)]{belousov2017f}
Boris Belousov and Jan Peters.
\newblock f-divergence constrained policy improvement.
\newblock \emph{arXiv preprint arXiv:1801.00056}, 2017.

\bibitem[Bertsekas(1999)]{Bertsekas99}
D.~P. Bertsekas.
\newblock \emph{Nonlinear Programming}.
\newblock Athena Scientific, Belmont, MA, second edition, 1999.

\bibitem[Bertsekas and Tsitsiklis(1996)]{bertsekas1996neuro}
Dimitri~P Bertsekas and John~N Tsitsiklis.
\newblock \emph{Neuro-dynamic programming}, volume~5.
\newblock Athena Scientific Belmont, MA, 1996.

\bibitem[Bertsekas et~al.(1995)Bertsekas, Bertsekas, Bertsekas, and
  Bertsekas]{bertsekas1995dynamic}
Dimitri~P Bertsekas, Dimitri~P Bertsekas, Dimitri~P Bertsekas, and Dimitri~P
  Bertsekas.
\newblock \emph{Dynamic programming and optimal control}, volume~1.
\newblock Athena scientific Belmont, MA, 1995.

\bibitem[Borwein and Lewis(2010)]{borwein2010convex}
Jonathan Borwein and Adrian~S Lewis.
\newblock \emph{Convex analysis and nonlinear optimization: theory and
  examples}.
\newblock Springer Science \& Business Media, 2010.

\bibitem[Boyd and Vandenberghe(2004)]{boyd2004convex}
Stephen Boyd and Lieven Vandenberghe.
\newblock \emph{Convex optimization}.
\newblock Cambridge university press, 2004.

\bibitem[Chen and Wang(2016)]{chen2016stochastic}
Yichen Chen and Mengdi Wang.
\newblock Stochastic primal-dual methods and sample complexity of reinforcement
  learning.
\newblock \emph{arXiv preprint arXiv:1612.02516}, 2016.

\bibitem[Chen et~al.(2018)Chen, Li, and Wang]{chen2018scalable}
Yichen Chen, Lihong Li, and Mengdi Wang.
\newblock Scalable bilinear $\pi$ learning using state and action features.
\newblock \emph{arXiv preprint arXiv:1804.10328}, 2018.

\bibitem[Dai et~al.(2016)Dai, He, Pan, Boots, and Song]{dai2016learning}
Bo~Dai, Niao He, Yunpeng Pan, Byron Boots, and Le~Song.
\newblock Learning from conditional distributions via dual embeddings.
\newblock \emph{arXiv preprint arXiv:1607.04579}, 2016.

\bibitem[Dai et~al.(2017)Dai, Shaw, He, Li, and Song]{dai2017boosting}
Bo~Dai, Albert Shaw, Niao He, Lihong Li, and Le~Song.
\newblock Boosting the actor with dual critic.
\newblock \emph{arXiv preprint arXiv:1712.10282}, 2017.

\bibitem[Dai et~al.(2019)Dai, Liu, Dai, He, Gretton, Song, and
  Schuurmans]{dai2019exponential}
Bo~Dai, Zhen Liu, Hanjun Dai, Niao He, Arthur Gretton, Le~Song, and Dale
  Schuurmans.
\newblock Exponential family estimation via adversarial dynamics embedding.
\newblock \emph{arXiv preprint arXiv:1904.12083}, 2019.

\bibitem[De~Farias and Van~Roy(2003)]{de2003linear}
Daniela~Pucci De~Farias and Benjamin Van~Roy.
\newblock The linear programming approach to approximate dynamic programming.
\newblock \emph{Operations research}, 51\penalty0 (6):\penalty0 850--865, 2003.

\bibitem[Denardo(1970)]{denardo1970linear}
Eric~V Denardo.
\newblock On linear programming in a markov decision problem.
\newblock \emph{Management Science}, 16\penalty0 (5):\penalty0 281--288, 1970.

\bibitem[Du et~al.(2017)Du, Chen, Li, Xiao, and Zhou]{du2017stochastic}
Simon~S Du, Jianshu Chen, Lihong Li, Lin Xiao, and Dengyong Zhou.
\newblock Stochastic variance reduction methods for policy evaluation.
\newblock In \emph{Proceedings of the 34th International Conference on Machine
  Learning-Volume 70}, pages 1049--1058. JMLR. org, 2017.

\bibitem[Ekeland and Temam(1999)]{EkeTem99}
Ivar Ekeland and Roger Temam.
\newblock \emph{Convex analysis and variational problems}, volume~28.
\newblock Siam, 1999.

\bibitem[Fu et~al.(2017)Fu, Luo, and Levine]{fu2017learning}
Justin Fu, Katie Luo, and Sergey Levine.
\newblock Learning robust rewards with adversarial inverse reinforcement
  learning.
\newblock \emph{arXiv preprint arXiv:1710.11248}, 2017.

\bibitem[Funk et~al.(2011)Funk, Westreich, Wiesen, St{\"u}rmer, Brookhart, and
  Davidian]{funk2011doubly}
Michele~Jonsson Funk, Daniel Westreich, Chris Wiesen, Til St{\"u}rmer, M~Alan
  Brookhart, and Marie Davidian.
\newblock Doubly robust estimation of causal effects.
\newblock \emph{American journal of epidemiology}, 173\penalty0 (7):\penalty0
  761--767, 2011.

\bibitem[Ghasemipour et~al.(2019)Ghasemipour, Zemel, and
  Gu]{ghasemipour2019divergence}
Seyed Kamyar~Seyed Ghasemipour, Richard Zemel, and Shixiang Gu.
\newblock A divergence minimization perspective on imitation learning methods.
\newblock \emph{arXiv preprint arXiv:1911.02256}, 2019.

\bibitem[Hasselt et~al.(2016)Hasselt, Guez, and Silver]{van2016deep}
H.~Van Hasselt, A.~Guez, and D.~Silver.
\newblock Deep reinforcement learning with double {Q}-learning.
\newblock In \emph{AAAI}, volume~16, pages 2094--2100, 2016.

\bibitem[Ho and Ermon(2016)]{ho2016generative}
Jonathan Ho and Stefano Ermon.
\newblock Generative adversarial imitation learning.
\newblock In \emph{Advances in neural information processing systems}, pages
  4565--4573, 2016.

\bibitem[Jiang and Li(2015)]{jiang2015doubly}
Nan Jiang and Lihong Li.
\newblock Doubly robust off-policy value evaluation for reinforcement learning.
\newblock \emph{arXiv preprint arXiv:1511.03722}, 2015.

\bibitem[Kallus and Uehara(2019{\natexlab{a}})]{kallus2019double}
Nathan Kallus and Masatoshi Uehara.
\newblock Double reinforcement learning for efficient off-policy evaluation in
  markov decision processes.
\newblock \emph{arXiv preprint arXiv:1908.08526}, 2019{\natexlab{a}}.

\bibitem[Kallus and Uehara(2019{\natexlab{b}})]{kallus2019efficiently}
Nathan Kallus and Masatoshi Uehara.
\newblock Efficiently breaking the curse of horizon: Double reinforcement
  learning in infinite-horizon processes.
\newblock \emph{arXiv preprint arXiv:1909.05850}, 2019{\natexlab{b}}.

\bibitem[Ke et~al.(2019)Ke, Barnes, Sun, Lee, Choudhury, and
  Srinivasa]{ke2019imitation}
Liyiming Ke, Matt Barnes, Wen Sun, Gilwoo Lee, Sanjiban Choudhury, and
  Siddhartha Srinivasa.
\newblock Imitation learning as $ f $-divergence minimization.
\newblock \emph{arXiv preprint arXiv:1905.12888}, 2019.

\bibitem[Kostrikov et~al.(2018)Kostrikov, Agrawal, Dwibedi, Levine, and
  Tompson]{kostrikov2018discriminator}
Ilya Kostrikov, Kumar~Krishna Agrawal, Debidatta Dwibedi, Sergey Levine, and
  Jonathan Tompson.
\newblock Discriminator-actor-critic: Addressing sample inefficiency and reward
  bias in adversarial imitation learning.
\newblock \emph{arXiv preprint arXiv:1809.02925}, 2018.

\bibitem[Kostrikov et~al.(2019)Kostrikov, Nachum, and
  Tompson]{kostrikov2019imitation}
Ilya Kostrikov, Ofir Nachum, and Jonathan Tompson.
\newblock Imitation learning via off-policy distribution matching, 2019.

\bibitem[Liu et~al.(2015)Liu, Liu, Ghavamzadeh, Mahadevan, and
  Petrik]{LiuLiuGhaMahetal15}
Bo~Liu, Ji~Liu, Mohammad Ghavamzadeh, Sridhar Mahadevan, and Marek Petrik.
\newblock Finite-sample analysis of proximal gradient td algorithms.
\newblock In \emph{Proc. The 31st Conf. Uncertainty in Artificial Intelligence,
  Amsterdam, Netherlands}, 2015.

\bibitem[Manne(1960)]{manne1960linear}
Alan~S Manne.
\newblock Linear programming and sequential decisions.
\newblock \emph{Management Science}, 6\penalty0 (3):\penalty0 259--267, 1960.

\bibitem[Mnih et~al.(2015)Mnih, Kavukcuoglu, Silver, Rusu, Veness, Bellemare,
  Graves, Riedmiller, Fidjeland, Ostrovski, et~al.]{mnih2015human}
Volodymyr Mnih, Koray Kavukcuoglu, David Silver, Andrei~A Rusu, Joel Veness,
  Marc~G Bellemare, Alex Graves, Martin Riedmiller, Andreas~K Fidjeland, Georg
  Ostrovski, et~al.
\newblock Human-level control through deep reinforcement learning.
\newblock \emph{Nature}, 518\penalty0 (7540):\penalty0 529, 2015.

\bibitem[Nachum et~al.(2019{\natexlab{a}})Nachum, Chow, Dai, and
  Li]{nachum2019dualdice}
Ofir Nachum, Yinlam Chow, Bo~Dai, and Lihong Li.
\newblock Dualdice: Behavior-agnostic estimation of discounted stationary
  distribution corrections.
\newblock \emph{arXiv preprint arXiv:1906.04733}, 2019{\natexlab{a}}.

\bibitem[Nachum et~al.(2019{\natexlab{b}})Nachum, Dai, Kostrikov, Chow, Li, and
  Schuurmans]{algae}
Ofir Nachum, Bo~Dai, Ilya Kostrikov, Yinlam Chow, Lihong Li, and Dale
  Schuurmans.
\newblock Algaedice: Policy gradient from arbitrary experience,
  2019{\natexlab{b}}.

\bibitem[Neu et~al.(2017)Neu, Jonsson, and G{\'o}mez]{neu2017unified}
Gergely Neu, Anders Jonsson, and Vicen{\c{c}} G{\'o}mez.
\newblock A unified view of entropy-regularized markov decision processes.
\newblock \emph{arXiv preprint arXiv:1705.07798}, 2017.

\bibitem[Nowozin et~al.(2016)Nowozin, Cseke, and Tomioka]{NowCseTom16}
Sebastian Nowozin, Botond Cseke, and Ryota Tomioka.
\newblock f-gan: Training generative neural samplers using variational
  divergence minimization.
\newblock \emph{arXiv preprint arXiv:1606.00709}, 2016.

\bibitem[Peng et~al.(2019)Peng, Kumar, Zhang, and
  Levine]{peng2019advantageweighted}
Xue~Bin Peng, Aviral Kumar, Grace Zhang, and Sergey Levine.
\newblock Advantage-weighted regression: Simple and scalable off-policy
  reinforcement learning, 2019.

\bibitem[Peters et~al.(2010)Peters, Mulling, and Altun]{peters2010relative}
Jan Peters, Katharina Mulling, and Yasemin Altun.
\newblock Relative entropy policy search.
\newblock In \emph{Twenty-Fourth AAAI Conference on Artificial Intelligence},
  2010.

\bibitem[Precup et~al.(2001)Precup, Sutton, and Dasgupta]{Precup01OP}
D.~Precup, R.~Sutton, and S.~Dasgupta.
\newblock Off-policy temporal difference learning with function approximation.
\newblock In \emph{Proceedings of the 18th International Conference on Machine
  Learning}, pages 417--424, 2001.

\bibitem[Puterman(1994)]{puterman1994markov}
Martin~L Puterman.
\newblock Markov decision processes: Discrete stochastic dynamic programming.
\newblock 1994.

\bibitem[Rockafellar(1970)]{rockafellar1970convex}
R~Tyrrell Rockafellar.
\newblock \emph{Convex analysis}, volume~28.
\newblock Princeton university press, 1970.

\bibitem[Scherrer(2010)]{scherrer2010should}
Bruno Scherrer.
\newblock Should one compute the temporal difference fix point or minimize the
  bellman residual? the unified oblique projection view.
\newblock \emph{arXiv preprint arXiv:1011.4362}, 2010.

\bibitem[Song et~al.(2019)Song, Abdolmaleki, Springenberg, Clark, Soyer, Rae,
  Noury, Ahuja, Liu, Tirumala, et~al.]{song2019v}
H~Francis Song, Abbas Abdolmaleki, Jost~Tobias Springenberg, Aidan Clark,
  Hubert Soyer, Jack~W Rae, Seb Noury, Arun Ahuja, Siqi Liu, Dhruva Tirumala,
  et~al.
\newblock V-mpo: On-policy maximum a posteriori policy optimization for
  discrete and continuous control.
\newblock \emph{arXiv preprint arXiv:1909.12238}, 2019.

\bibitem[Sutton et~al.(2014)Sutton, Mahmood, Precup, and
  Hasselt]{sutton2014new}
Rich Sutton, Ashique~Rupam Mahmood, Doina Precup, and Hado Hasselt.
\newblock A new q $(\lambda)$ with interim forward view and monte carlo
  equivalence.
\newblock In \emph{International Conference on Machine Learning}, pages
  568--576, 2014.

\bibitem[Sutton(1996)]{sutton1996generalization}
Richard~S Sutton.
\newblock Generalization in reinforcement learning: Successful examples using
  sparse coarse coding.
\newblock In \emph{Advances in neural information processing systems}, pages
  1038--1044, 1996.

\bibitem[Sutton and Barto(1998)]{sutton1998introduction}
Richard~S Sutton and Andrew~G Barto.
\newblock \emph{Introduction to reinforcement learning}, volume 135.
\newblock 1998.

\bibitem[Sutton et~al.(2000)Sutton, McAllester, Singh, and
  Mansour]{sutton2000policy}
Richard~S Sutton, David~A McAllester, Satinder~P Singh, and Yishay Mansour.
\newblock Policy gradient methods for reinforcement learning with function
  approximation.
\newblock In \emph{Advances in neural information processing systems}, pages
  1057--1063, 2000.

\bibitem[Sutton et~al.(2008)Sutton, Maei, and Szepesvári]{SutMaeSze08}
Richard~S. Sutton, Hamid~R. Maei, and Csaba Szepesvári.
\newblock A convergent $o(n)$ temporal-difference algorithm for off-policy
  learning with linear function approximation.
\newblock In D.~Koller, D.~Schuurmans, Y.~Bengio, and L.~Bottou, editors,
  \emph{Advances in Neural Information Processing Systems 21}, pages
  1609--1616. 2008.

\bibitem[Sutton et~al.(2009)Sutton, Maei, Precup, Bhatnagar, Silver,
  Szepesv{\'a}ri, and Wiewiora]{SutMaePreBhaetal09}
Richard~S Sutton, Hamid~Reza Maei, Doina Precup, Shalabh Bhatnagar, David
  Silver, Csaba Szepesv{\'a}ri, and Eric Wiewiora.
\newblock Fast gradient-descent methods for temporal-difference learning with
  linear function approximation.
\newblock In \emph{Proceedings of the 26th Annual International Conference on
  Machine Learning}, pages 993--1000. ACM, 2009.

\bibitem[Sutton et~al.(2016)Sutton, Mahmood, and White]{sutton2016emphatic}
Richard~S Sutton, A~Rupam Mahmood, and Martha White.
\newblock An emphatic approach to the problem of off-policy temporal-difference
  learning.
\newblock \emph{The Journal of Machine Learning Research}, 17\penalty0
  (1):\penalty0 2603--2631, 2016.

\bibitem[Tang et~al.(2019)Tang, Feng, Li, Zhou, and Liu]{tang2019doubly}
Ziyang Tang, Yihao Feng, Lihong Li, Dengyong Zhou, and Qiang Liu.
\newblock Doubly robust bias reduction in infinite horizon off-policy
  estimation.
\newblock \emph{arXiv preprint arXiv:1910.07186}, 2019.

\bibitem[Uehara and Jiang(2019)]{uehara2019minimax}
Masatoshi Uehara and Nan Jiang.
\newblock Minimax weight and q-function learning for off-policy evaluation.
\newblock \emph{arXiv preprint arXiv:1910.12809}, 2019.

\bibitem[Veinott(2005)]{veinott}
Arthur F.~Jr Veinott.
\newblock Conjugate and lagrangian duality for convex programs.
\newblock \url{https://web.stanford.edu/class/msande361/handouts/nlpdual.pdf},
  2005.
\newblock Accessed: 2019-11-14.

\bibitem[Wainwright and Jordan(2003)]{WaiJor03}
M.~J. Wainwright and M.~I. Jordan.
\newblock Graphical models, exponential families, and variational inference.
\newblock Technical Report 649, UC Berkeley, Department of Statistics,
  September 2003.

\bibitem[Wang(2017)]{wang2017randomized}
Mengdi Wang.
\newblock Randomized linear programming solves the discounted markov decision
  problem in nearly-linear (sometimes sublinear) running time.
\newblock \emph{arXiv preprint arXiv:1704.01869}, 2017.

\bibitem[Wang et~al.(2007)Wang, Bowling, Lizotte, and Schuurmans]{wang2007dual}
Tao Wang, Michael Bowling, Daniel Lizotte, and Dale Schuurmans.
\newblock Dual representations for dynamic programming.
\newblock 2007.

\bibitem[Wang et~al.(2015)Wang, Schaul, Hessel, Van~Hasselt, Lanctot, and
  De~Freitas]{wang2015dueling}
Ziyu Wang, Tom Schaul, Matteo Hessel, Hado Van~Hasselt, Marc Lanctot, and Nando
  De~Freitas.
\newblock Dueling network architectures for deep reinforcement learning.
\newblock \emph{arXiv preprint arXiv:1511.06581}, 2015.

\bibitem[Watkins(1989)]{watkins1989learning}
Christopher John Cornish~Hellaby Watkins.
\newblock Learning from delayed rewards.
\newblock 1989.

\bibitem[Zhang et~al.(2020)Zhang, Dai, Lihong, and
  Schuurmans]{zhang2020gendice}
Ruiyi Zhang, Bo~Dai, Li~Lihong, and Dale Schuurmans.
\newblock {GenDICE}: Generalized offline estimation of stationary values.
\newblock 2020.
\newblock Preprint.

\end{thebibliography}

\end{document}